\documentclass[letterpaper]{article} 
\usepackage{aaai23}  
\usepackage{times}  
\usepackage{helvet}  
\usepackage{courier}  
\usepackage[hyphens]{url}  
\usepackage{graphicx} 
\usepackage{natbib}  
\usepackage{caption} 
\usepackage{amsmath,amsfonts,amsthm,bm}
\usepackage{algorithm}
\usepackage{algorithmic}
\usepackage{dsfont}
\usepackage{mathtools}

\usepackage[labelfont={},labelformat=simple]{subcaption}

\usepackage{bibentry}

\title{Causal Explanation for Reinforcement Learning: \\
Quantifying State and Temporal Importance}

\author {
    Xiaoxiao Wang,\textsuperscript{\rm 1}
    Fanyu Meng, \textsuperscript{\rm 1}
    Xin Liu, \textsuperscript{\rm 1}
    Zhaodan Kong, \textsuperscript{\rm 1}
    Xin Chen \textsuperscript{\rm 2}
}
\affiliations {
    \textsuperscript{\rm 1} University of California, Davis\\
    \textsuperscript{\rm 2} Georgia Institute of Technology\\
    \{xxwa, fymeng, xinliu, zdkong\}@ucdavis.edu, xinchen@gatech.edu 
}

\usepackage{bibentry}

\begin{document}

\nocopyright
\maketitle

\begin{abstract}
Explainability plays an increasingly important role in machine learning. Furthermore, humans view the world through a causal lens and thus prefer causal explanations over associational ones. Therefore, in this paper, we develop a causal explanation mechanism that quantifies the causal importance of states on actions and such importance over time. We also demonstrate the advantages of our mechanism over state-of-the-art associational methods in terms of RL policy explanation through a series of simulation studies, including crop irrigation, Blackjack, collision avoidance, and lunar lander.
\end{abstract}

\section{Introduction}

Reinforcement learning (RL) is  increasingly being considered in domains with significant social and safety implications such as healthcare, transportation, and finance. This growing societal-scale impact has raised a set of concerns, including trust, bias, and explainability. For example, can we explain how an RL agent arrives at a certain decision? When a policy performs well, can we explain why? These concerns mainly arise from two factors. First, many popular RL algorithms, particularly deep RL, utilize neural networks, which are essentially black boxes with their inner workings being opaque not only to lay persons but also to data scientists. Second, RL is a trial-and-error learning algorithm in which an agent tries to find a policy that minimizes a long-term reward by repeatedly interacting with its environment. Temporal information such as relationships between states at different time instances plays a key role in RL and subsequently adds another layer of complexity compared to supervised learning.

The field of explainable RL (XRL), a sub-field of explainable AI (XAI), aims to partially address these concerns by providing explanations as to why an RL agent arrives at a particular conclusion or action. While still in its infancy, XRL has made good progress over the past few years, particularly by taking advantage of existing XAI methods  \cite{puiutta2020explainable,heuillet2021explainability,wells2021explainable}. For instance, inspired by the saliency map method~\cite{simonyan2014deep} in supervised learning which explains image classifiers by highlighting ``important" pixels in terms of classifying images, some XRL methods attempt to explain the decisions made by an RL agent by generating maps that highlight ``important" state features~\cite{iyer2018transparency,greydanus2018visualizing,mott2019towards}. However, there exist at least two major limitations in state-of-the-art XRL methods. First, the majority of them take an \textit{associational} perspective. For instance, the aforementioned studies quantify the ``importance'' of a feature by calculating the correlation between the state feature and an action. Since it is well known that ``correlation doesn't imply causation'' \cite{pearl2009causality}, it is possible that features with a high correlation may not necessarily be the real ``cause'' of the action,  resulting in a misleading explanation that can lead to user skepticism and possibly even rejection of the RL system. Second, \textit{temporal} information is not generally considered. Temporal effects, such as the interaction between states and actions over time, which as mentioned previously is essential in RL, are not taken into account.

\begin{figure}[ht]
\centering
\includegraphics[width=\linewidth]{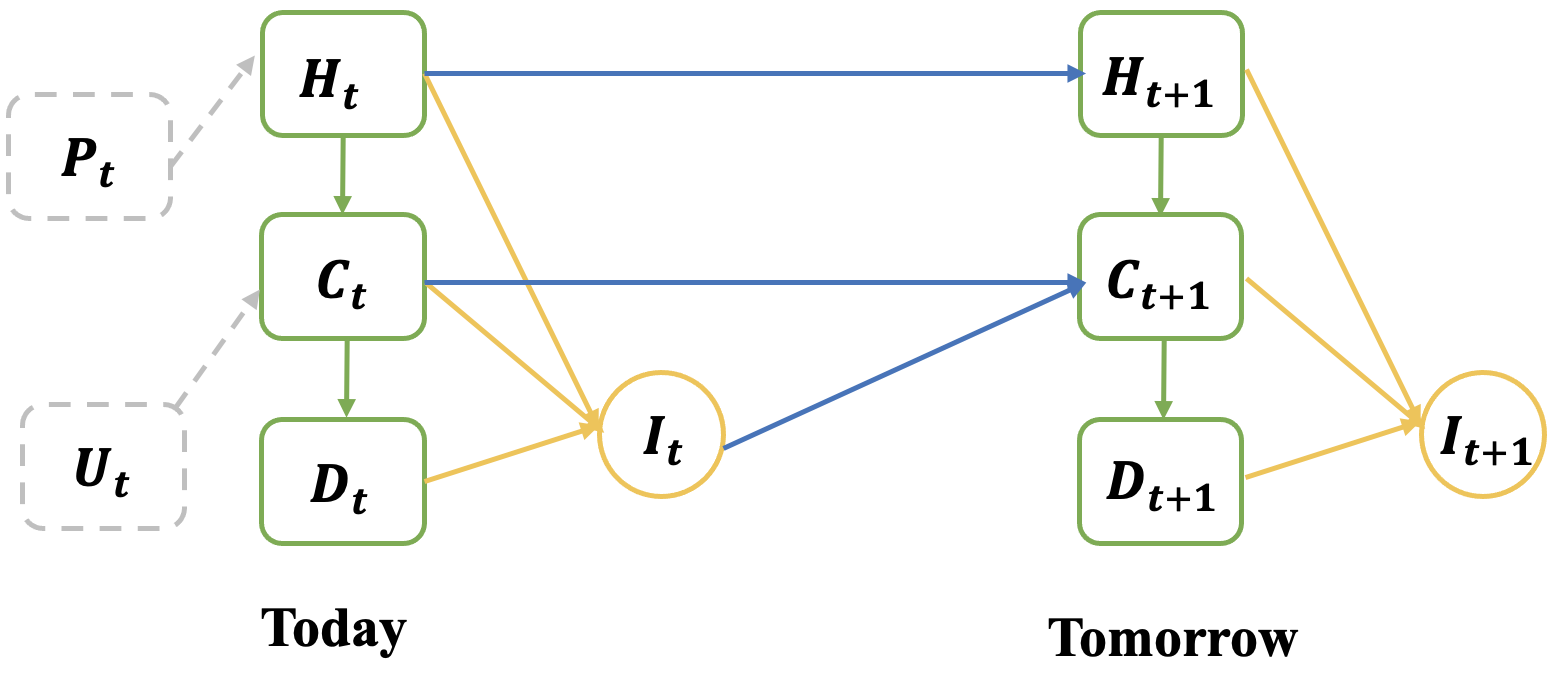}
\caption{Causal graph of the crop irrigation problem. Endogenous and exogenous states are denoted by dashed and solid rectangles, respectively, while actions are denoted by circles. More details about causal graphs can be found in the \textbf{Preliminaries} section.}
\label{gp:agri}
\end{figure}

In this paper, we propose a \textit{causal} XRL mechanism. Specifically, we explain an RL policy by incorporating a causal model that we have about the relationship between states and actions. To best illustrate the key features of our XRL mechanism, we use a concrete crop irrigation problem as an example, as shown in Fig.~\ref{gp:agri} (more details can be found in the \textbf{Evaluation} section). In this problem, an RL policy $\pi$ controls the amount of irrigation water ($I_t$) based on the following endogenous (observed) state variables: humidity ($H_t$), crop weight ($C_t$), and radiation ($D_t$). Its goal is to maximize the crop yield during harvest.
Crop growth is also affected by some other features, including the observed precipitation ($P_t$) and other exogenous (unobserved) variables $U_t$. 
To explain why policy $\pi$ arrives at a particular action $I_t$ at the current state, our XRL method quantifies the \textit{causal} importance of each state feature, such as $H_t$, in the context of this action $I_t$ via \textit{counterfactual reasoning} \cite{byrne2019counterfactuals,miller2019explanation}, i.e., by calculating how the action would have changed if the feature had been different.

Our proposed XRL mechanism addresses the aforementioned limitations as follows. First, our method can generate inherently causal explanations. To be more specific, in essence, importance measures used in associational methods can only capture \textit{direct} effects while our causal importance measures capture \textit{total} causal effects. For example, for the state feature $H_t$, our method can account for two causal chains: the direct effect chain $H_t\to I_t$ and the indirect effect chain $H_t\to C_t \to I_t$, while associational methods only consider the former. Second, our method can quantify the temporal effect between actions and states, such as the effect of today's humidity $H_t$ on tomorrow's irrigation $I_{t+1}$. In contrast, associational methods, such as saliency map~\cite{greydanus2018visualizing}, cannot measure how previous state features can affect the current action because their models only formulate the relationship between state and action in one time step and ignore  temporal relations. To the best of our knowledge, our XRL mechanism is the \textit{first work} that explains RL policies by causally explaining their actions based on causal state and temporal importance. 
It has been studied that humans are more receptive to a contrastive explanation, i.e., humans answer a ``Why X?" question through the answer to the often only implied-counterfactual ``Why not Y instead?" \cite{hilton2007causal,miller2019explanation}. Because our causal explanations are based on contrastive samples, users may find our explanations more intuitive.

\section{Related Work}

\paragraph{Explainable RL (XRL)} Based on how an XRL algorithm generates its explanation, we can categorize existing XRL methods into state-based, reward-based, and global surrogate explanations \cite{puiutta2020explainable,heuillet2021explainability,wells2021explainable}. State-based methods explain an action by highlighting state features that are important in terms of generating the action \cite{greydanus2018visualizing,puri2019explain}. Reward-based methods generally apply reward decomposition and identify the sub-rewards that contribute the most to decision making~\cite{juozapaitis2019explainable}. Global surrogate methods generally approximate the original RL policy with a simpler and transparent (also called intrinsically explainable) surrogate model, such as decision trees, and then generate explanations with the surrogate model~\cite{verma2018programmatically}. In the context of state-based methods, there are generally two ways to quantify feature importance: (i) gradient-based methods, such as simple gradient~\cite{simonyan2013deep} and integrated gradients~\cite{sundararajan2017axiomatic}, and (ii) sensitivity-based methods, such as LIME~\cite{ribeiro2016should} and SHAP~\cite{lundberg2017unified}. Our work belongs to the category of state-based methods. However, instead of using associations to calculate importance, a method generally used in existing state-based methods, our method adopts a causal perspective. The benefits of such a causal approach have been discussed in the \textbf{Introduction} section.

\paragraph{Causal Explanation} Causality has already been utilized in XAI, mainly in supervised learning settings. Most existing studies quantify feature importance by either using Granger causality~\cite{schwab2019cxplain} and  average or individual causal effect metric~\cite{chattopadhyay2019neural} or by applying random valued interventions~\cite{datta2016algorithmic}. Two recent studies \cite{madumal2020explainable} and \cite{olson2021counterfactual} are both focused on causal explanations in an RL setting.  Compared with \cite{madumal2020explainable}, the main difference is that we provide a different type of explanation. Our method involves finding an importance vector that quantifies the impact of each state feature, while \cite{madumal2020explainable} provides a causal chain starting from the action. We also demonstrate the ability of our approach to provide temporal importance explanations that can capture the impact of a state feature or action on the future state or action. This aspect has been discussed in the crop irrigation experiment in Section \ref{sec:crop_simulation}. Additionally, we construct  structural causal models(SCM) differently. While the action is modeled as an edge in the SCM in the paper \cite{madumal2020explainable}, our method formulates  the action as a vertex in the SCM model, allowing us to quantify the state feature impact on action. As for \cite{olson2021counterfactual}, our approach is unique in that it can calculate the temporal importance of a state, which is not achievable by their method. Furthermore, we have provided a value-based importance definition of Q-value that differs from their method. Another significant difference between our approach and \cite{olson2021counterfactual} is the underlying assumption. Our method takes into account intra-state relations, which are ignored in Olson's work. Neglecting intra-state causality is more likely to result in an invalid state after the intervention, leading to inaccurate estimates of importance. Therefore, our approach considers the causal relationships between state features to provide a more accurate and comprehensive explanation of the problem.

\section{Preliminaries}
We introduce the notations used throughout the paper. We use capital letters such as $X$ to denote a random variable and small letters such as $x$ for its value. Bold letters such as $\mathbf{X}$ denote a vector of random variables and  superscripts such as  $\mathbf{X}^{(i)}$ denote its $i$-th element. Calligraphic letters such as $\mathcal{X}$ denote sets.  For a given natural number $n$, $[n]$ denotes the set $\{1,2,\cdots, n\}$.

\paragraph{Causal Graph and Skeleton}
Causal graphs are probabilistic graphical models that define data-generating processes~\cite{pearl2009causality}. 
Each vertex of the graph represents a variable.  Given a set of variables $\mathcal{V} = \{V_i, i \in [n]\}$, a directed edge from a variable $V_j$ to $V_i$ denotes that $V_i$ responds to changes in $V_j$ when all other variables are held constant.
Variables connected to $V_i$ through directed edges are defined as the parents of $V_i$, or ``direct causes of $V_i$," and the set of all such variables is denoted by $\mathcal{P}a_{i}$.  The skeleton of a causal graph is defined as the topology of the graph. The skeleton can be obtained using background knowledge or learned using causal discovery algorithms, such as the classical constraint-based PC algorithm~\cite{spirtes2000causation} and those based on linear non-Gaussian models~\cite{shimizu2006linear}. In this work, we assume the skeleton is given.

\paragraph{SCM}  
In a causal graph, we can define the value of each variable $V_i$ as a function of its parents and  exogenous variables. Formally, we have the following definition of SCM:
let $\mathcal{V} =\{ V_i, i \in [n]\}$ be a set of endogenous(observed)  variables and $\mathcal{U} =\{ U_i, i \in [n]\}$ be a set of exogenous(unobserved) variables. A SCM~\cite{pearl2009causality} is defined as a set of structural equations in the form of
\begin{equation}
\label{eq:scm}
    V_i = f_i(\mathcal{P}a_{i},U_i),  \mathcal{P}a_i \subset \mathcal{V},  U_i \subset \mathcal{U}, i \in [n],
\end{equation}
where function $f_i$ represents a causal mechanism that determines the value of $V_i$ using its parents and the exogenous variables.
\paragraph{Intervention and Do-operation} 
SCM can be used for causal interventions, denoted by the $do(\cdot) $ operator. $do(V_i=v)$ means setting the value of $V_i$ to a constant $v$ regardless of its structural equation in the SCM, i.e., ignoring the edges into the vertex $V_i$.
Note that the do-operation differs from the conditioning operation in statistics. Conditioning on a variable implies  information about its parent variables due to correlation.

\paragraph{Counterfactual Reasoning}
Counterfactual reasoning allows us to answer ``what if" questions. For example, assume that the state is $X_t = x$ and the action is $A_t=a$. We are interested in knowing what would have happened if the state had been at a different value $x'$. This implies a counterfactual question~\cite{pearl2009causality}. The counterfactual outcome of $A_t$ can be represented as $A_{t,X_t=x'}  \vert  X_t = x, A_t= a$. Given an SCM, we can perform counterfactual reasoning based on intervention through the following two steps: 
\begin{enumerate}
    \item Recover the value of exogenous variable $U$ as $u$ through the structural function $f$ and the values $X_t = x$, $A_t = a$;
    \item Calculate the counterfactual outcome as $A_t  \vert do(X_t = x'), U=u$. More specifically, in SCM, we set up the value of $X_t$ to $x'$. Then we substitute all exogenous variable values to the right side of the functions and get the counterfactual outcome $A_t$.
\end{enumerate}

\paragraph{MDP and RL}
An infinite-horizon Markov Decision Process (MDP) is a tuple $(\mathcal{S},\mathcal{A},P,R)$ , where $\mathcal{S} \in \mathbb{R}^m$ and $\mathcal{A} \in \mathbb{R}$ are finite sets of states and actions, $P (\mathbf{s}, a, \mathbf{s}')$ is the probability of transitioning from state $\mathbf{s}$ to state $\mathbf{s}'$ after taking action $a$, and $R(\mathbf{s},a)$ is the reward for taking $a$ in $\mathbf{s}$. An RL policy $\pi$ returns an action to take at state $\mathbf{s}$, and its associated Q-function, $Q_\pi (\mathbf{s}, a)$, provides the expected infinite-horizon $\gamma$-discounted cumulative reward for taking action $a$ at state $\mathbf{s}$ and following $\pi$ thereafter.

\section{Problem Formulation} 

Our focus is on policy explainability, and we assume that the policy $\pi$ and its associated Q-function, $Q_\pi (\mathbf{s}, a)$, are given. Note that the policy may or may not be optimal.
We require a dataset containing trajectories of the agent interacting with the MDP using the policy $\pi$. A single trajectory consists of a sequence of $(\mathbf{s}, a, r, \mathbf{s'})$ tuples.
Additionally, We  assume that the skeleton of the causal graph, such as the one shown in  Fig.~\ref{gp:agri} for the crop irrigation problem, is known. We do not assume that the SCM, more specifically its structural functions, is given. We assume the additive noise for the SCM but not its linearity (discussed in Eq.~(\ref{eq:additive}) in Section~\ref{sec:importance-vector-for-state}). The {\bf goal} is to answer the question ``why does the policy $\pi$ select the current action $a$ at the current state $\mathbf{s}$?" We provide causal explanations for this question from two perspectives: state importance and temporal importance.
\paragraph{Importance vector for state}
The first aspect of our explanation is to use the important state feature to provide an explanation. Specifically, we seek to construct an {\bf importance vector} for the state, where each dimension measures the impact of the corresponding state feature on the action. 
 For instance, in the crop irrigation problem, we can answer the question ``why does the RL agent irrigate more water today?" by stating that ``the impact of humidity, crop weight,  and radiation on the current irrigation decision is quantified as  $[0.8,0.1,0.1]$ respectively.
Formally, we have the following definition of the importance vector for state explanation. Given state $\mathbf{s}_t$ and policy $\pi$, the importance of each feature of $\mathbf{s}_t$ for the current action $a_t$ is quantified as $\mathbf{w}_t$. The explanation is that the features in state $\mathbf{s}_t$ have causal importance $\mathbf{w}_t$ on policy $\pi$ to select action $a_t$ at  state $\mathbf{s}_t$. 

\paragraph{Temporal importance of action/state}
The second aspect of our explanation considers the temporal aspect of RL. Here, we measure how the actions and states in the past impact the current action. We can generalize the importance vector above to past states and actions. 
Formally, given state $\mathbf{s}_t$, policy $\pi$ and the history trajectory of the agent $\mathcal{H}_t := \{(\mathbf{s}_\tau, a_\tau), \tau \leq t\}$, we define the effect of a past action $a_\tau$ on the current action $a_t$ as $w^{a_\tau}_t$. Similarly, for a past state  $\mathbf{s}_{\tau}$, we define the temporal importance vector $\mathbf{w}_t^{\tau}$, in which each dimension measures the impact of the corresponding state feature at time step $\tau$ on current action $a_t$. Then we  use $w^{a_\tau}_t$ and $\mathbf{w}_t^{\tau}$ to quantify the impact of  past states and action.

\section{Explanation}

\subsection{Importance Vector for State}
\label{sec:importance-vector-for-state}
Our mechanism implements the following two steps to obtain the importance vector $\mathbf{w}_t$.

\begin{enumerate}
    \item Train  SCM structural functions between the states and actions using the data of historical trajectories of the RL agent;
    \item Compute the important vector by intervening in the SCM. 
\end{enumerate}
\begin{figure}[ht]
\centering
\includegraphics[width=\linewidth]{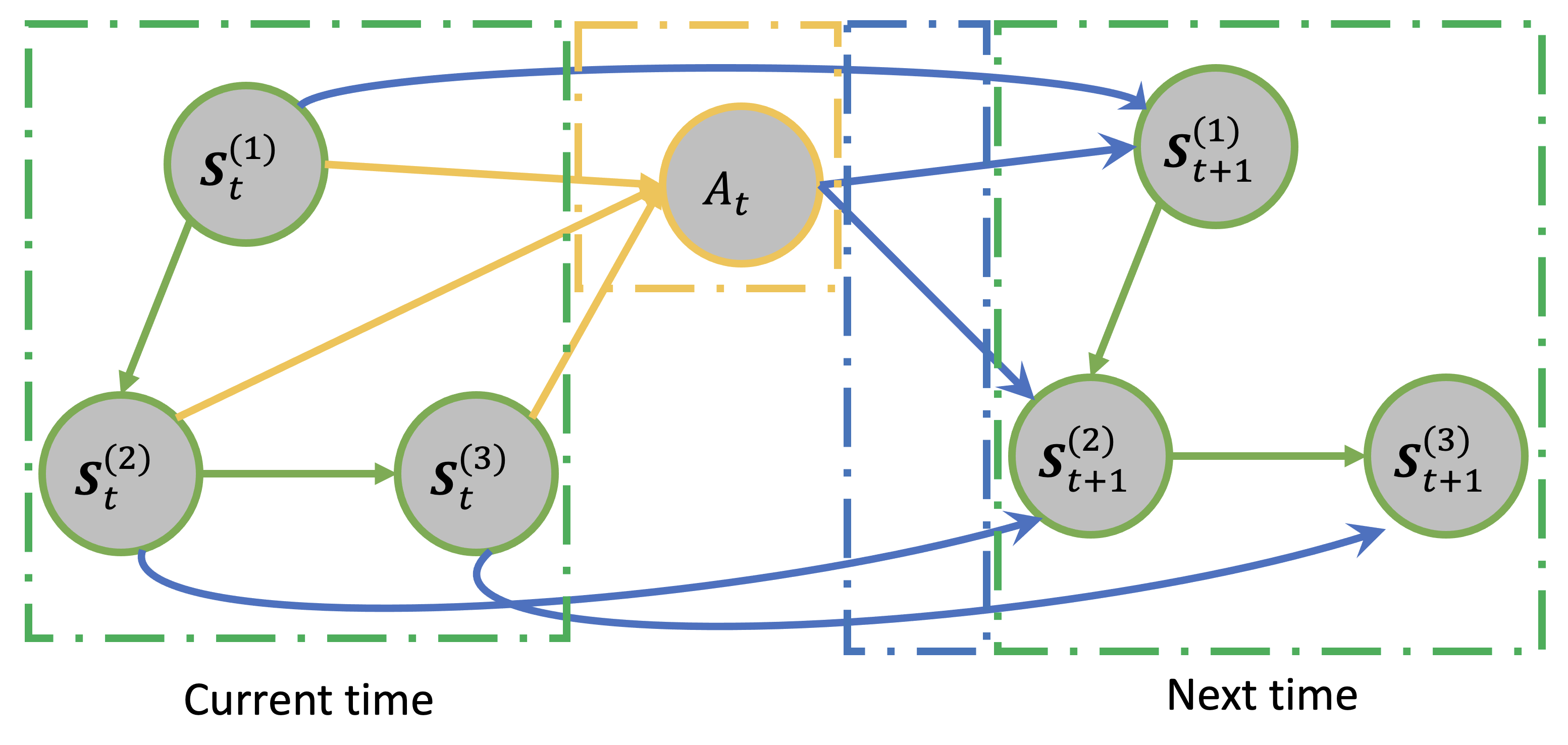} 
\caption{Example causal graph between the state and action. $\mathbf{S}_{t}^{(i)}$ is the $i$-th dimension of the  interested state $\mathbf{S}$ at time $t$. Each vertex also has a corresponding exogenous variable, which has no parent and its only child is the associated endogenous variable. Per causality conventions, the exogenous variables are omitted in the graph.}
\label{gp:causal_state_action}
\end{figure}
First, we notice that there are three types of causal relations between the states and actions: intra-state, policy-defined, and transition-defined relations. As shown in Fig.~\ref{gp:causal_state_action}, the green directed edges represent the intra-state relations, which are defined by the underlying causal mechanism. The orange edges describe the policy and represent how the state variables affect the action. The third type of relation shown as blue edges is the causal relationship between the states across different times. They represent the dynamics of the environment and depend on the transition probability $P(\mathbf{s}_t,a_t,\mathbf{s}_{t+1})$ in the MDP.

We assume that the intra-state and transition-defined causal relations are captured by the causal graph skeleton. For the policy-defined relations, we assume a general case where all state features are the causal parents of the action. In the causal graph, each edge defines a causal relation, and the vertex defines a variable $V$ with a causal structural function $f$. Then we only need to learn the causal structural functions between the vertices. To achieve this, we can learn each vertex's function separately. For a vertex $V_i$ and its parents  $\mathcal{P}a_i$, based on Eq.~(\ref{eq:scm}), we make an additive noise assumption to simplify the problem and formulate the function mapping between $V_i$ and $\mathcal{P}a_i$ as
\begin{equation}
\label{eq:additive}
   V_i = f_i(\mathcal{P}a_i) + U_i, 
\end{equation}
where $U_i$ is an exogenous variable. We note that the additive noise assumption is widely used in the causal discovery literature~\cite{hoyer2008nonlinear,peters2014causal}. We then use supervised learning to learn the function mapping among the vertices. Specifically, $f_a$ for action $a_t$ is defined as 
\[ A_t= f_a(\mathbf{S}_{t}^{(1)},\cdots,\mathbf{S}_{t}^{(m)}, U_{a}), \]
where $m$ is the dimension of the state, and $U_{a}$ is the exogenous variable for the actions.

For the state variables, we denote all exogenous variables as a vector $\mathbf{U_S}:=[U_1, \cdots, U_m]$ and learn the structural functions. Intuitively, the exogenous variables $U_{a}$ and $\mathbf{U_S}$ represent not only  random noise but also  hidden features or the stochasticity of the policy for the intra-state and policy-defined causal relations. For transition-defined relations, the exogenous variables can be regarded as the stochasticity in the environment.

\subsection{Action-based Importance} 
Given a state $\mathbf{s}_{t}$
and an action $a_t $, the importance vector $\mathbf{w}_t$ is calculated by applying intervention on the learned SCM. Based on the additive noise assumption, we  recover the values of the exogenous variables $\mathbf{U_s}$ and $U_a$ according to the value of $a_t$, $\mathbf{s_t}$ and the learned causal structural functions. Then we define $\mathbf{w}_t$ using the intervention operation (counterfactual reasoning). Specifically, we define the importance vector $\mathbf{w}_{t}=[\mathbf{w}_{t}^{(1)}, \cdots,\mathbf{w}_{t}^{(m)}]$ as 
\begin{equation}
\label{eq:impor_ve}
    \mathbf{w}_{t}^{(i)} = \frac{  \left\vert  \left( \left. A_{t, \mathbf{S}_{t}^{(i)}=\mathbf{s}_{t}^{(i)}+\delta} \right\vert \mathbf{S}_{t} = \mathbf{s}_{t}, A_t=a_t\right) - a_t  \right\vert }{\delta},
\end{equation}
where $ \vert  \cdot  \vert $ is a vector norm (e.g., absolute-value norm) and $\delta$ is a small perturbation value chosen according to the problem setting. 
The term $A_{t, \mathbf{S}_{t}^{(i)}=\mathbf{s}_{t}^{(i)}+\delta} \vert \mathbf{S}_{t} = \mathbf{s}_{t}, A_t=a_t$ represents the counterfactual outcome of $A_t$ if we set $\mathbf{S}_{t}^{(i)}=\mathbf{s}_{t}^{(i)}+\delta$. In our case, the value of the exogenous variables can be recovered using the additive noise assumption, so the value of $A_{t, \mathbf{S}_{t}^{(i)}=\mathbf{s}_{t}^{(i)}+\delta} \vert \mathbf{S}_{t} = \mathbf{s}_{t}, A_t=a_t$ can be determined. 
We interpret the result as that the features with a larger $\mathbf{w}_{t}^{(i)}$ have a more significant causal impact on  the agent's action $a_t$. 
Note that in the simulation, we average the importance from both positive and negative $\delta$ and return the average as the final score. The perturbation amount $\delta$ is a hyperparameter and should be selected according to each problem setting.

\subsection{Q-value-based Importance} 
While action-based importance can capture the causal impact of states on the change of the action, it may not capture the more subtle causal importance when the selected action does not change, especially when the action space is discrete. 
Specifically, $A_{t, \mathbf{S}_{t}^{(i)}=\mathbf{s}_{t}^{(i)}+\delta}  \vert \mathbf{S}_{t} = \mathbf{s}_{t}, A_t=a_t$ may not change after a perturbation of $\delta$, which will result in a $ \mathbf{w}_{t}^{(i)} = 0$. However, this is different from when there are no causal paths from feature $ \mathbf{S}_{t}^{(i)}$ to the action $A_t$, also resulting in a $ \mathbf{w}_{t}^{(i)} = 0$. Therefore, we also define Q-value-based importance as follows: 
\begin{equation}
{^Q}\mathbf{w}_{t}^{(i)}  = \frac{  \vert  Q_\pi^\text{perturb} - Q_\pi(\mathbf{s}_{t},a_t) \vert }{\delta},
\label{eq:impor_q}
\end{equation}
where $Q_\pi^\text{perturb} = Q_\pi(\mathbf{S}_{t,\mathbf{S}_{t}^{(i)}=\mathbf{s}_{t}^{(i)}+\delta}, A_{t,\mathbf{S}_{t}^{(i)}=\mathbf{s}_{t}^{(i)}+\delta}   \vert \mathbf{S}_{t} \!=\! \mathbf{s}_{t}, A_t\!=\!a_t)$.
In detail, we use counterfactual reasoning to compute the  counterfactual outcome of $A_t$ and $S_t$ after setting $\mathbf{S}_{t}^{(i)}=\mathbf{s}_{t}^{(i)}+\delta$ and then substituting them into $Q_{\pi}$ to evaluate the corresponding Q-value. 
Similar to the action-based importance, we account for both positive and negative importance in practice.
 See the Blackjack Section~\ref{sc:blackjack} in evaluation for a comparison between Eq.~(\ref{eq:impor_ve}) and Eq.~(\ref{eq:impor_q}) on an example with a discrete action space.

In most RL algorithms, Q-value critically impacts which actions to choose. Therefore, we consider Q-valued-based importance as explanations on the action through the Q-value. However, we note that the Q-value-based importance method sometimes cannot reflect which features the policy really depends on. Some features may contribute largely to the Q-value of all state-action pairs ({\{$Q(\mathbf{s}_t, 
a_t), a_t \in \mathcal{A} \}$}, but not to the decision making process - the action with the largest Q-value ($\arg\max_{a_t\in \mathcal{A}}Q(\mathbf{s}_t, a_t)$). In such cases, these features may have an equal impact on the Q-value regardless of the action. For example, in the crop irrigation problem, crop pests have an impact on the crop yield (Q-value) but don't impact the amount of irrigation water (the action). 
Some related simulations are shown in Appendix~\ref{sec:compareQ}. In summary, we suggest using the action-based importance method by default and the Q-value-based method as a supplement. 

\subsection{Temporal Importance and Cascading SCM}
Temporal importance allows us to quantify the impact of past states and actions on the current action. 
In RL, estimating of temporal effect is important because 
 policies are generally non-myopic, and  actions should affect all future states and actions. 
To measure the importance beyond the previous step, we define an extended causal model that includes state features and actions in the previous time step, as shown in Fig.~\ref{gp:agri}. In this model, the vertices in the graph are $\{\mathbf{S}_\tau, A_\tau\}_{\tau=1}^T$. For simplicity, we  assume the system is stationary, so the causal relations are stationary and do not change over time. Therefore, the structural functions are the same as those defined in Fig.~\ref{gp:causal_state_action}, i.e., the mechanism of an edge $(\mathbf{S}_\tau^{(i)}, \mathbf{S}_{\tau+1}^{(j)})$ will be the same as the edge $(\mathbf{S}_t^{(i)}, \mathbf{S}_{t+1}^{(j)})$. The extended causal model can be regarded as a cascade of multiple copies of the same module, where each module is similar to that in Fig.~\ref{gp:causal_state_action}. We can  estimate the effect of perturbing any features or actions at any step through intervention, and the effect will propagate through the modules to the final time step. We illustrate the temporal importance in the Blackjack experiment in Section~\ref{sc:blackjack}.

\subsection{Comparison with Associational Methods}
\label{sec:compare_assoicate}
In Eq.~(\ref{eq:impor_ve}), we define importance by applying intervention. If we change the $do$ action to the conditioning operation, we have the following definition, which is the same as the association-based saliency map method:
\begin{equation}
\label{eq:impor_s}
    {_{\text{sal}}}\mathbf{w}_{t}^{(i)}
    \!=\! \frac{  \left.\vert  A_{t}  \vert \mathbf{S}_{t}\!=\![\mathbf{s}_{t}^{(1)},\cdots\!,\mathbf{s}_{t}^{(i)}\!+\!\delta,\cdots\!, \mathbf{s}_{t}^{(m)}] - a_{t} \right\vert }{\delta}
\end{equation}

Associational models cannot perform individual-level counterfactual reasoning and hence cannot infer the counterfactual outcome after changing the value of one feature of the current state.
As pointed out by~\cite{pearl2009causality}, counterfactual reasoning can infer the specific property of the considered individual that is related to the exogenous variables, and then derives what would have happened if the agent had been in an alternative state. In our method, we use counterfactual reasoning to recover the environment at the current state and estimate how the action responds to the change in one of the state features. So our causal importance can capture more insights compared to the associational methods.

\begin{figure}[htb]
\centering
\begin{minipage}{0.2\linewidth}
  \includegraphics[width=\linewidth]{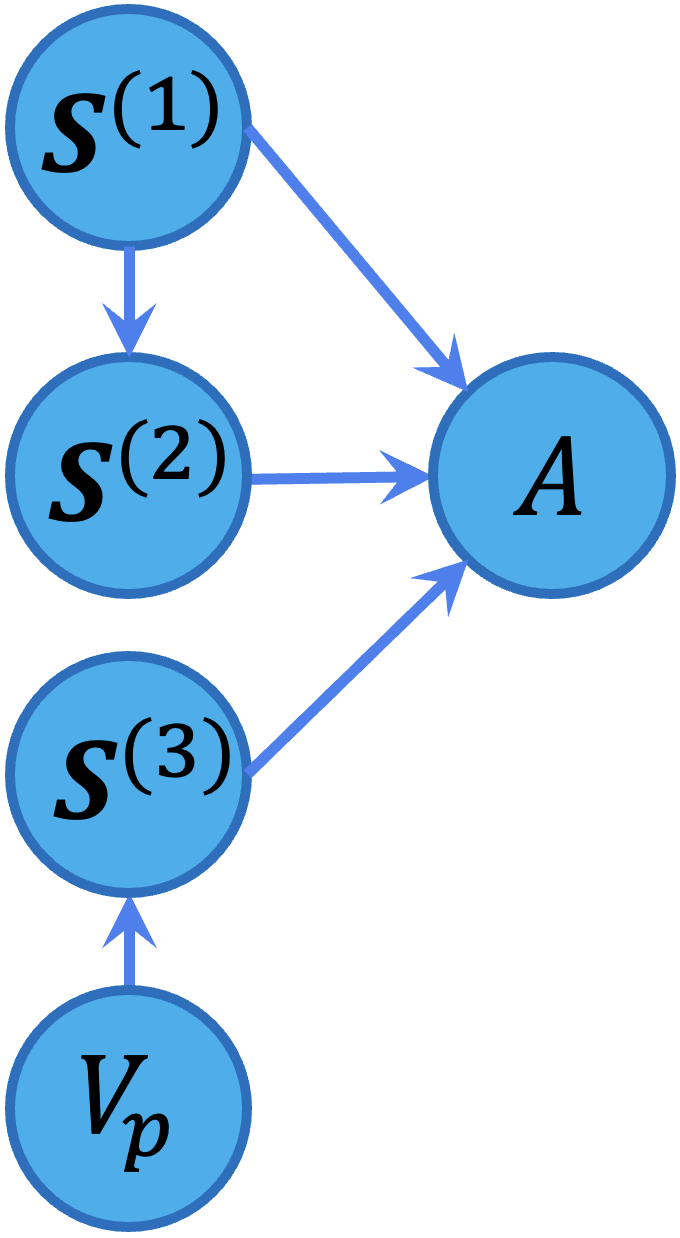}
\end{minipage}
\begin{minipage}{0.7\linewidth}
\begin{align}
\label{eq:th}
  \begin{cases}
    \mathbf{S}^{(1)}&=\!U_1 \\
    \mathbf{S}^{(2)}&=\!c_{12} \mathbf{S}^{(1)}\! + \!U_2 \\
    \mathbf{S}^{(3)}&=\!c_pV_{p} \!+\! U_3 \\
        A&=\!c_1\mathbf{S}^{(1)}\!+\!c_{2}(\mathbf{S}^{(2)})^2\!+\! c_3\mathbf{S}^{(3)}\! + \!U_a \\
    V_p&=\!U_{p}
    \end{cases} 
\end{align}
\end{minipage}
\caption{Example of a one-step MDP}
\label{gp:thexp}
\end{figure}

In Fig.~\ref{gp:thexp}, we use a one-step MDP toy example to demonstrate the difference. Omitting the time step subscript in the notation, we assume the policy is defined on the state space $\mathbf{S} =[\mathbf{S}^{(1)}, \mathbf{S}^{(2)}, \mathbf{S}^{(3)}]$. An observed variable $V_p$ is a causal parent of  $\mathbf{S}^{(3)}$ but is not defined in the state space. We define the ground truth of the state and policy as Eq.~(\ref{eq:th}), where $c_1, c_2, c_3, c_{12}, c_p$ are constant parameters and $U_a, U_1, U_2, U_3, U_{p}$ are exogenous variables. We use a  linear SCM to show the difference between the two methods. We do not assume the SCM to have linear dependencies.

We assume that both the associational method saliency map and our causal method can learn the ground truth functions. Given a state $\mathbf{s}$, the importance vectors using the two methods are compared in Table~\ref{tb:score}. We notice that, for $\mathbf{s}^{(1)}$, our method can capture the effect of $\mathbf{s}^{(1)}$ through two causal chains $\mathbf{S}^{(1)} \!\to\! A$ and $\mathbf{S}^{(1)} \!\to\! \mathbf{S}^{(2)} \!\to\! A$, while the saliency map method captures only $\mathbf{S}^{(1)} \!\to\! A$. Our causal method considers the fact that a change in $\mathbf{S}^{(1)}$ will result in a change of $\mathbf{S}^{(2)}$ and thus additionally influence the action $A$. The non-direct paths are also meaningful in explanation and should be considered in  measuring  the importance of $\mathbf{S}^{(1)}$. However, they are ignored in the saliency map method. The causal importance vector for $\mathbf{s}^{(1)}$ also considers the effect of $u_2$, which is recovered through counterfactual reasoning. This makes the causal-based importance specific to the current state. Additionally, our method can calculate the effect of $V_p$ on the action $A$, which can not be achieved by the  associational method saliency map.

\begin{table}[htp]	
\centering
\caption{Importance vector on the environment in Fig.~\ref{gp:thexp} using  our method and the saliency map method.}
\label{tb:score}
\begin{tabular}{ | l | l | l |}
 \hline
  & Our method & Saliency map \\
 \hline
 $\mathbf{s}^{(1)}$  & $c_1+c_2c_{12}(c_{12}(2\mathbf{s}^{(1)} +\delta)+2u_2)$  & $c_1$  \\
 $\mathbf{s}^{(2)}$  & $c_2(2\mathbf{s}^{(2)} + \delta)$  & $c_2(2\mathbf{s}^{(2)} + \delta)$  \\
 $\mathbf{s}^{(3)}$  & $c_3$  & $c_3$  \\
 $v_p$  & $c_pc_3$  & N/A  \\
\hline
\end{tabular}
\end{table}
We also note that for features $\mathbf{s}^{(2)}$ and $\mathbf{s}^{(3)}$, the two methods obtain the same result. In cases where a state feature is (1) not a causal parent of other features, (2) the policy is deterministic, and (3) there are no exogenous variables, our method is equivalent to the saliency-style approach. However, these conditions may not be common in RL. In general, there are causal relations among state features, such as the chess positions in the game of chess, the state features [position, velocity, acceleration] in a self-driving problem, and the state features [radiation, temperature, humidity] in a greenhouse control problem.

\section{Evaluation}

We test our causal explanation framework in three toy environments: crop irrigation (Section \ref{sec:crop_simulation}), collision avoidance (Section \ref{sec:bang_bang}), and Blackjack (Section \ref{sc:blackjack}). We also conduct experiments on Lunar Lander, which is a more sophisticated RL environment (Appendix \ref{app:lunar_lander}). For each experiment, the system dynamics, policy, training details, and perturbation values used can be found in Appendix \ref{app:addition_experiments_and_details}.

\subsection{Crop Irrigation Problem}
\label{sec:crop_simulation}

\begin{figure}[htbp]
  \centering
  \includegraphics[width=0.73\linewidth]{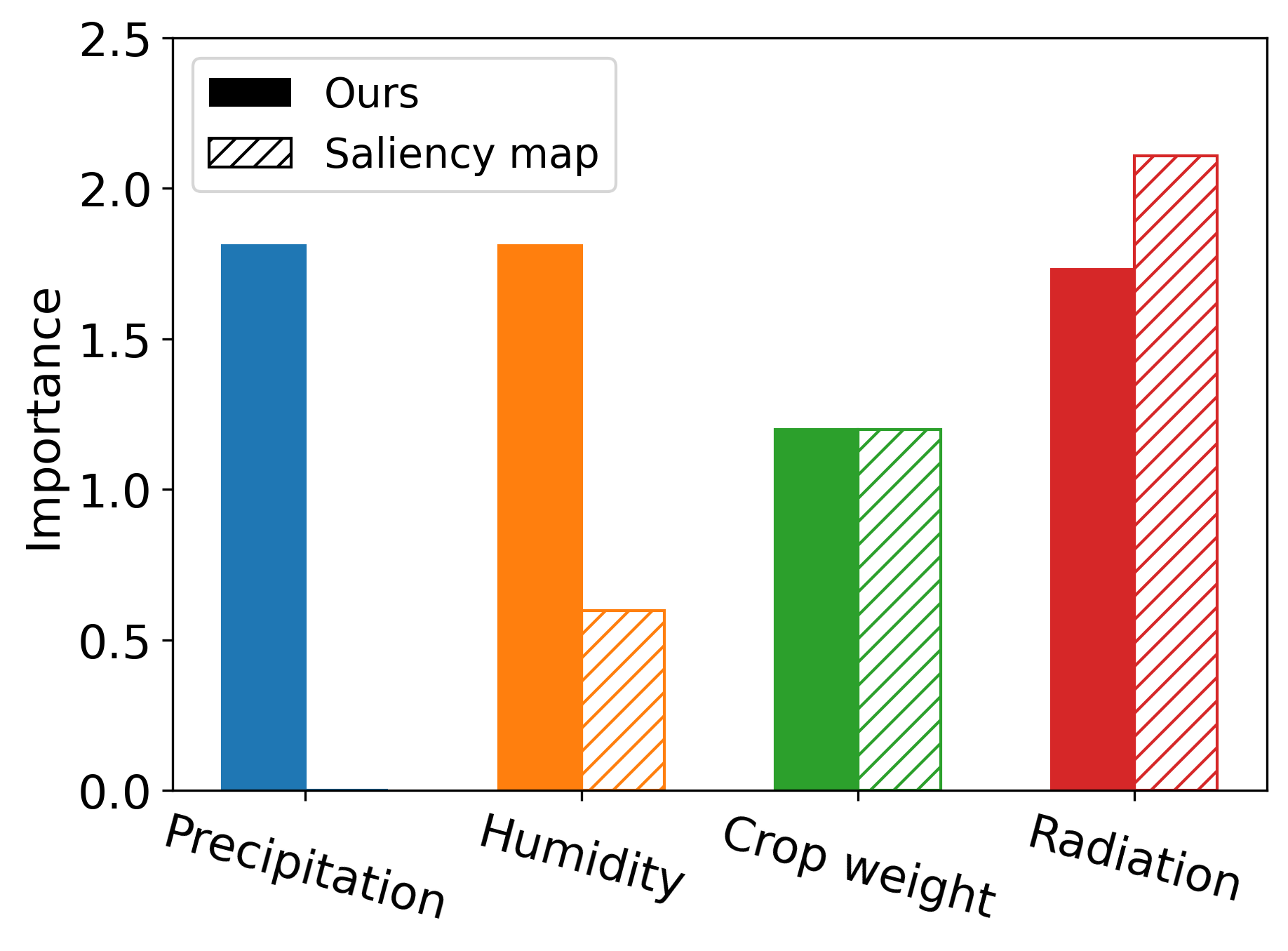}
  \caption{The importance vector for the crop irrigation problem.}
  \label{gp:agri_compare}
\end{figure}

We show the results of our explanation algorithm for the crop irrigation problem.  We assume a simplified environment dynamic based on agriculture models~\cite{williams1989epic}. The growth of the plant at each step is determined by the state features humidity ($H_t$), crop weight ($C_t$), and radiation ($D_t$). The policy controls the amount of water to irrigate each day. Intuitively, it irrigates more when the crop weight is high, and less when the crop weight is low.  Details about the environment dynamics and policy are described in Appendix \ref{app:crop}. We use Fig. \ref{gp:agri} as the causal skeleton and apply a neural network to learn the structural equations. Fig.~\ref{gp:agri_compare} shows the importance vector of the state for a given environment $[P_t=0.07,H_t=0.12, C_t=0.44, D_t=0.70]$ and its corresponding action $I_t=0.67$. First, we notice that our method can estimate the importance of the feature precipitation($P_t)$, which is not defined in the state space of the policy. Second, in estimating the causal importance of $H_t$, our method can estimate the effect of $H_t\to C_t \to I_t$, which results in  higher importance compared to the saliency map method. Since an intervention on $H_t$ can induce a change in $C_t$, causing the action to change more drastically. This effect cannot be measured without a causal model. The same  applies to the feature $D_t$. The full trajectory and the importance vector at each time step can be found in Fig.~\ref{gp:grop_compare_time} in Appendix \ref{app:crop}.

The causality-based action influence model~\cite{madumal2020explainable} can find a causal chain $I_t\to\ C_t \to\,$\texttt{CropYield} and  provide the explanation as ``the agent takes current action to increase $C_t$ at this step, which aims to increase the eventual crop yield.'' This explanation only provides the information that $C_t$ is an important factor in the decision-making for the current action  but can't quantify it. Moreover, this explanation can't provide  information for other state features, such as $H_t$  and $D_t$ which are also measured in our importance vector.

\subsection{Collision Avoidance Problem}
\label{sec:bang_bang}
We use a collision avoidance problem to further illustrate that our causal method can find a more meaningful importance vector than saliency map, i.e., which state feature is more impactful to decision-making.

\begin{figure}[htbp]
\centering
\captionsetup{justification=centering}
\begin{subfigure}[c]{\linewidth}
  \centering
  \includegraphics[width=0.7\linewidth]{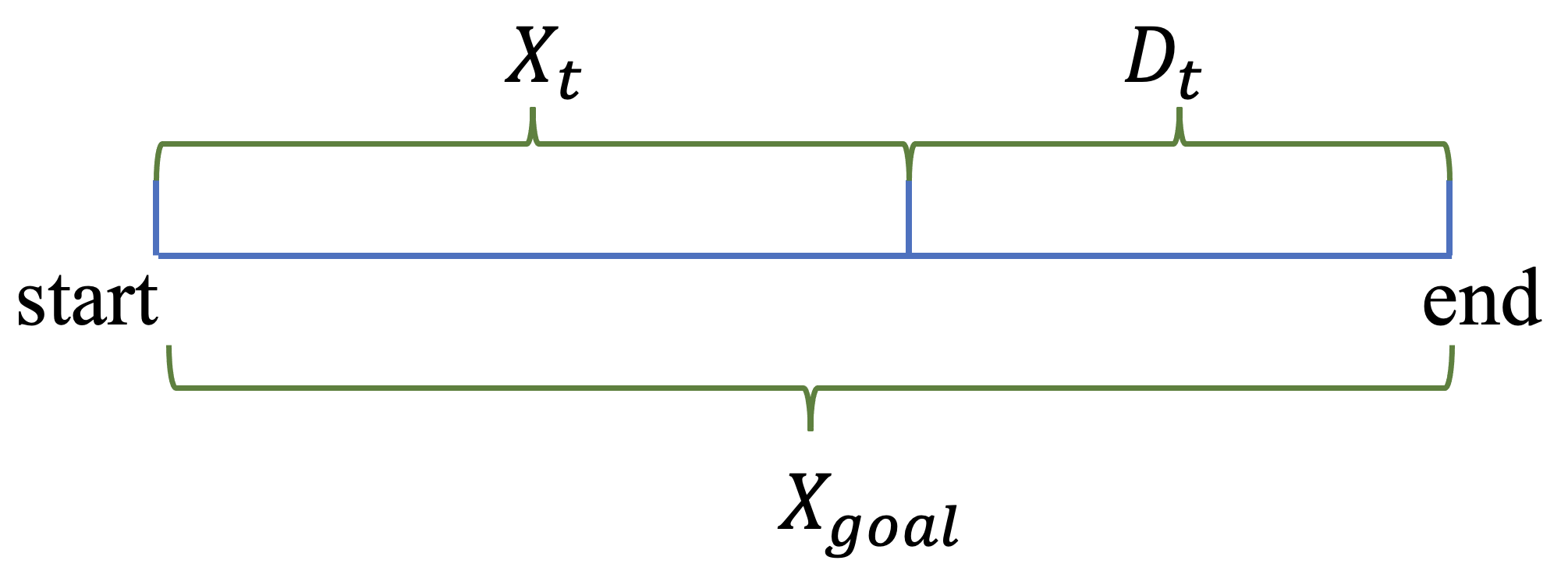}
  \caption{The state definition}
  \label{gp:bang}
\end{subfigure}
\begin{subfigure}[c]{\linewidth} 
  \centering
  \includegraphics[width=0.7\linewidth]{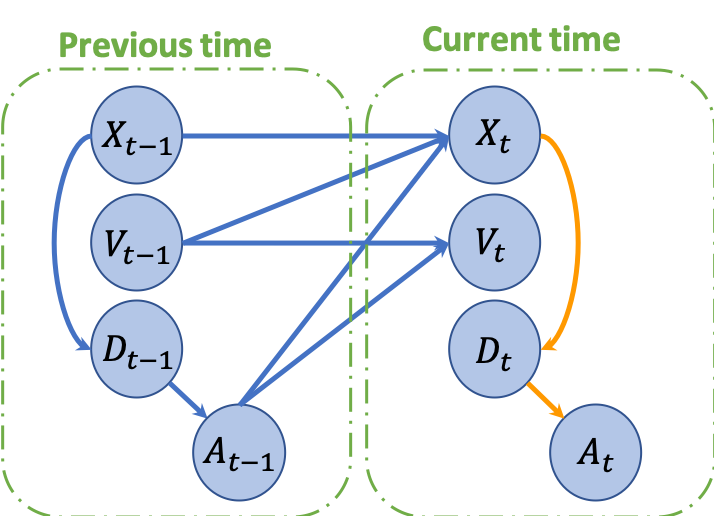}
  \caption{The causal graph of states and actions}
  \label{gp:bang_caual}
\end{subfigure}
\caption{The collision avoidance problem and its corresponding SCM skeleton.}
\end{figure}

Fig.~\ref{gp:bang} shows the state definition for this problem. A car with zero initial velocity travels from the start point to an endpoint over a distance of $X_\textrm{goal}$. The system is controlled in a discrete-time-slot manner and we assume acceleration of the car is constant within each time step.
The state $\mathbf{S}_t$ includes the distance from the start $X_t$, the distance to the end $D_t$, and the velocity $V_t$ of the car, i.e., $\mathbf{S}_t := [V_t, X_t, D_t]$, where $V_t \leq v_{\max}$ and $v_{\max}$ is the maximum speed of the car. The action $A_t$ is the car's acceleration, which is bounded  $\vert A_t \vert \leq e_{\max}$. We assume the acceleration of the car is constant within each time step. More detailed settings are described in the simulation section in the supplementary materials. The objective is to find a policy $\pi$ to minimize the traveling time under the condition that the final velocity is zero at the endpoint (collision avoidance).

An RL agent learns the following {\bf optimal} control policy for this avoidance problem, which is also known as the bang-bang control  (optimal under certain technical conditions) \cite{bryson1975applied}:

\begin{equation}
\label{eq:bangpolicy}
  A_{t} =
    \begin{cases}
      e_{\max} & \text{if } D_t \leq {v_{\max}^2} / ({2e_{\max}})\\
      -e_{\max} & \text{otherwise}
    \end{cases}       
\end{equation}
Intuitively, this policy accelerates as much as possible until reaching the critical point defined above. Then it will decelerate until reaching the goal. 

We use Fig. \ref{gp:bang_caual} as the SCM skeleton and use linear regression to learn the structural equations as the entire dynamics are linear. The detail about the system dynamics is described in the appendix. 

\begin{figure}[htbp]
\centering
\captionsetup{justification=centering}
\begin{subfigure}{\linewidth}
\centering
  \includegraphics[width=0.7\linewidth]{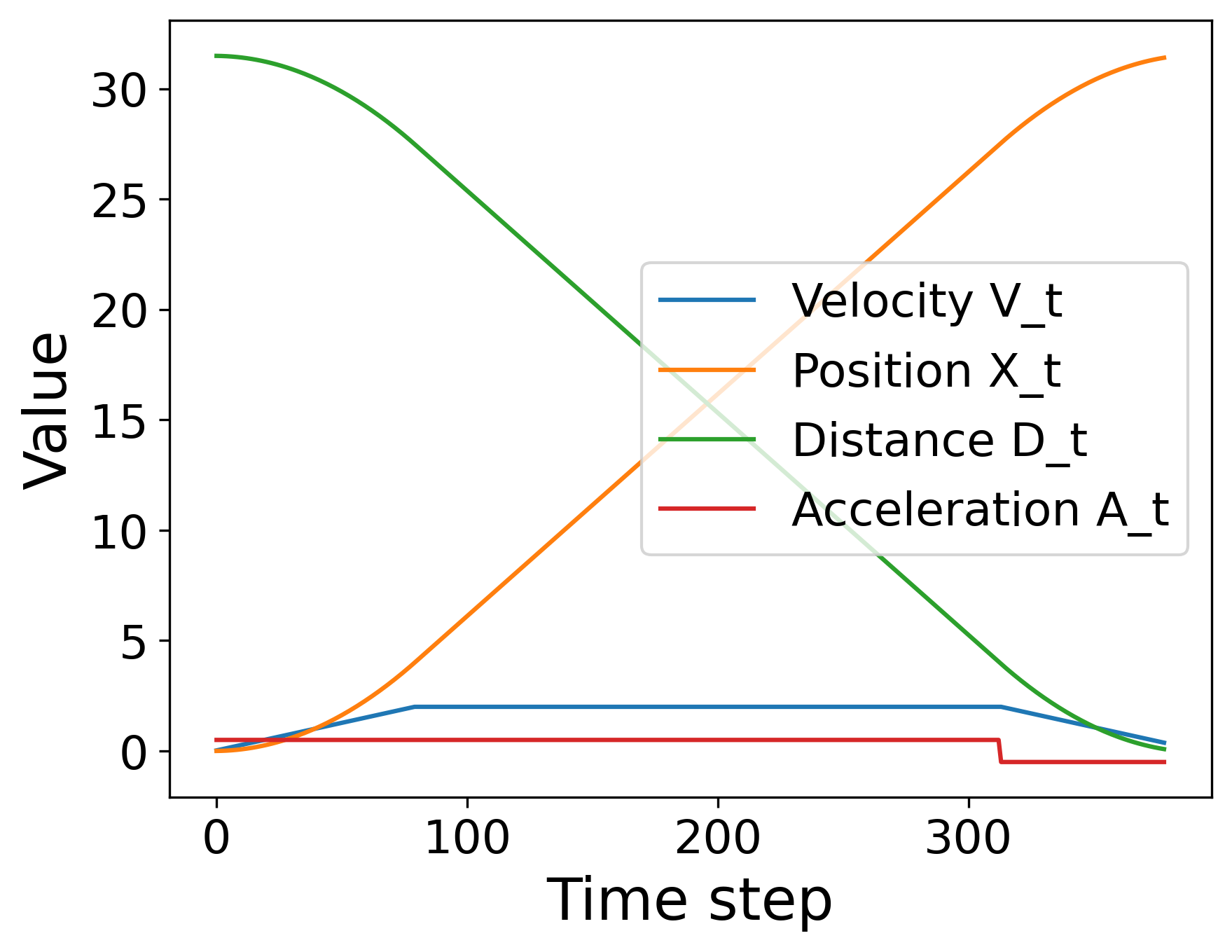}
  \caption{A trajectory of using bang-bang control on the collision avoidance problem.}
  \label{gp:bang_bang_trajectory}
\end{subfigure}
\begin{subfigure}{\linewidth}
\centering
  \includegraphics[width=0.7\linewidth]{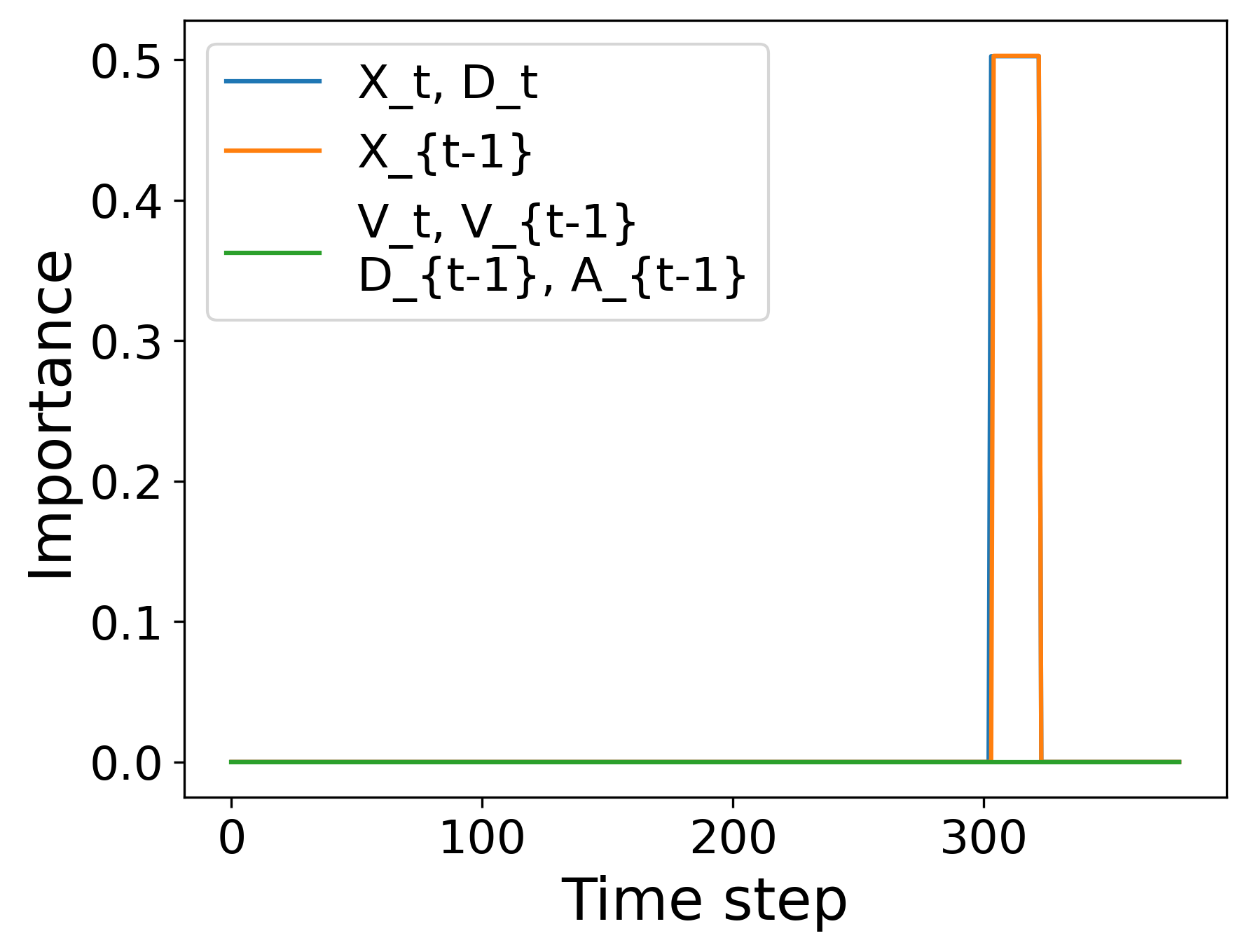}
  \caption{The result importance from our algorithm on the trajectory in Fig.~\ref{gp:bang_bang_trajectory}.}
  \label{gp:bang_importance}
\end{subfigure}
\caption{Trajectory and importance on the collision avoidance problem}
\label{fig:synthetic_scenarios}
\end{figure}

Fig.~\ref{gp:bang_bang_trajectory} shows a trajectory under the policy bang-bang control and Fig.~\ref{gp:bang_importance} shows its corresponding causal importance results. The importance of $V_t, A_{t-1}, D_{t-1}, V_{t-1}$ are zero throughout the time history, and those of $X_t, D_t, X_{t-1}$ have peak importance of $[0.502, 0.502, 0.502]$ , respectively,  between time step 303-322, during which the car changes the direction of acceleration to avoid hitting the obstacle. The importance curves of $X_t$, $D_t$, and $X_{t-1}$ have the same shape, but that of $X_{t-1}$ is off by one time step, corresponding to their time step subscript. If we were to use the associational saliency method~\cite{greydanus2018visualizing} $X_t$ would have a constant zero importance since the action is solely determined by the feature $D_t$. In comparison, our method can find non-zero importance through the edge $X_t \to D_t$. It is reasonable that $X_t$ causally affects $A_t$, because, in the physical world, the path length $X_t$ is the cause of the measurement of the distance to the end $D_t$. Although in Eq.~(\ref{eq:bangpolicy}) the action $A_t$ is only decided by $D_t$, the source cause of the change in $D_t$ is $X_t$. We can only obtain such information through a causal model, not an associational one.

\subsection{Blackjack}
\label{sc:blackjack}

We test our explanation mechanism on a simplified game of Blackjack. The state is defined as [\texttt{hand}, \texttt{ace}, \texttt{dealer}], where \texttt{hand} represents the sum of current cards in hand, \texttt{ace} represents if the player has a usable ace (an ace that can either be a 1 or an 11), and \texttt{dealer}, is the value of the dealer's shown card. There are two possible actions: to \texttt{draw} a new card or to \texttt{stick} and end the game. We use an on-policy Monte-Carlo control~\cite{sutton2018reinforcement} agent to test our mechanism. Since the problem dynamic is non-linear, we use a neural network to learn each structural equation. Fig. \ref{gp:blackjack_scm} shows the skeleton of the SCM. More details about the rules of the game are explained in Appendix \ref{app:black}.
Note that in Blackjack, the exogenous variable $U_i$ of some features can be interpreted as the stochasticity or the ``luck" during the input trajectory. e.g., $U_{\texttt{hand}, t}$ corresponds to the value of the card drawn at step $t$ if the previous action is \texttt{draw}.

\begin{figure}[htbp]
  \centering
  \includegraphics[width=0.7\linewidth]{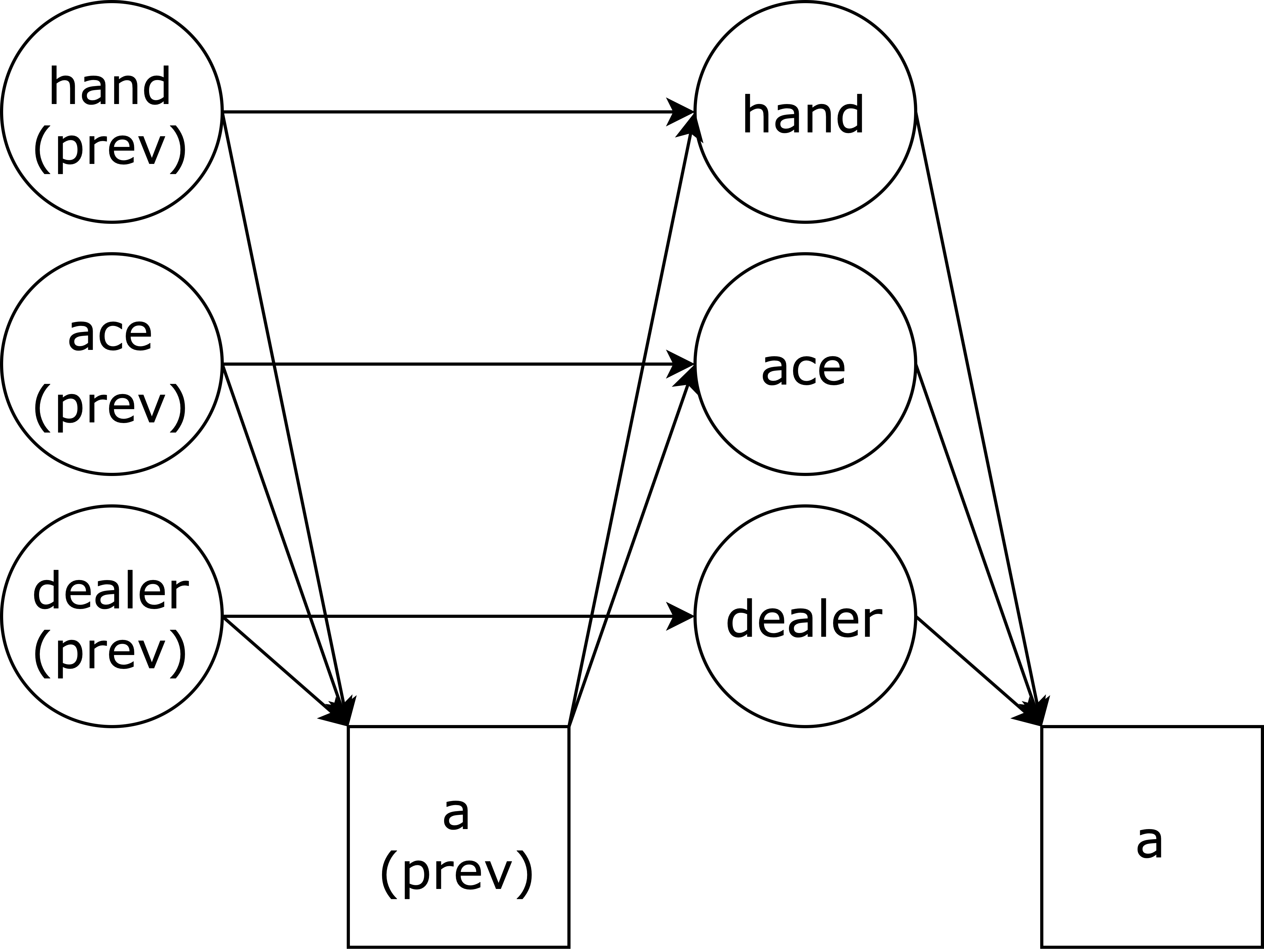}
  \caption{The skeleton of the Blackjack SCM.}
  \label{gp:blackjack_scm}
\end{figure}

\begin{figure*}[htbp]
  \centering
  \includegraphics[width=0.88\textwidth]{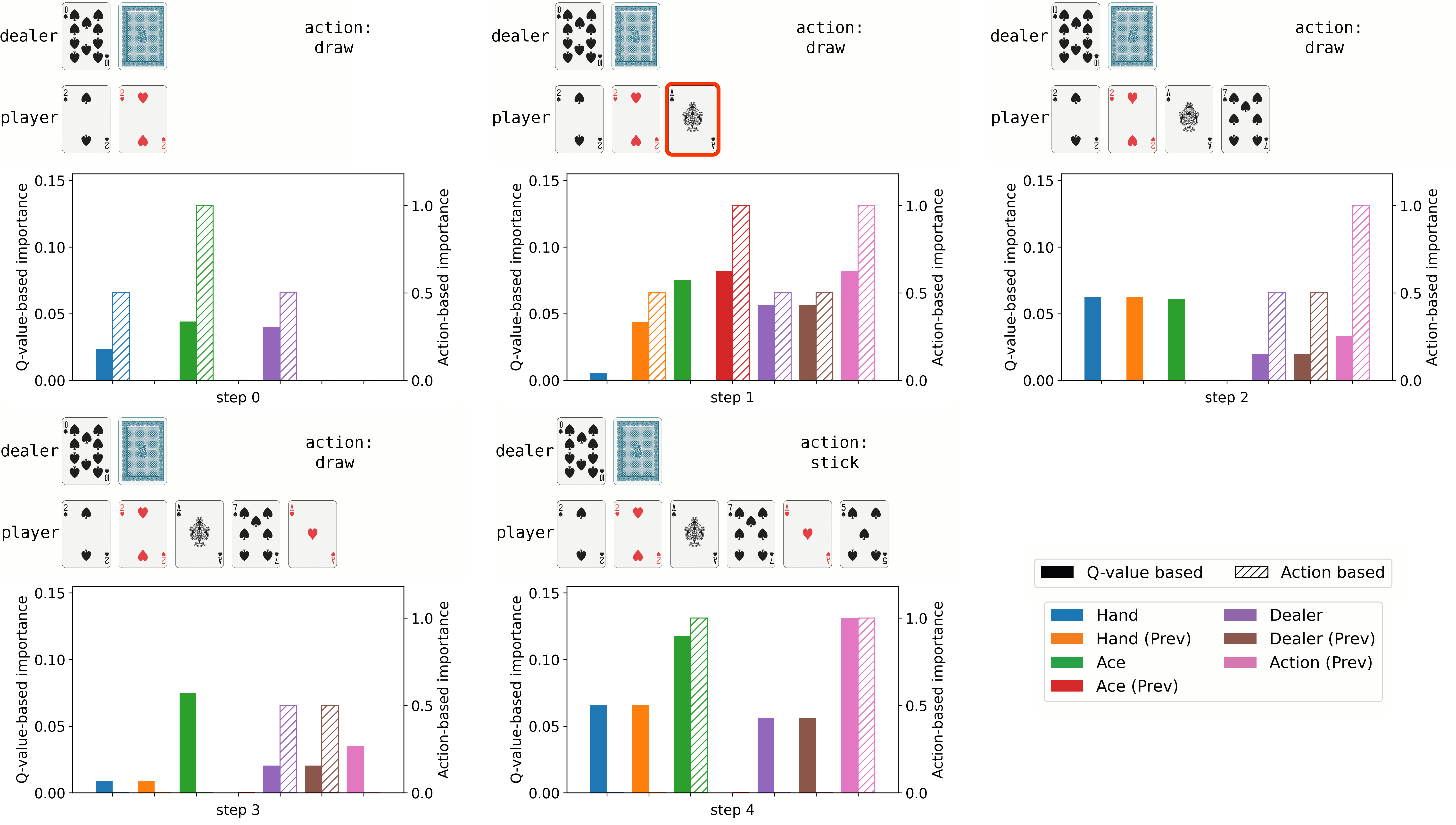}
  \caption{A trajectory of a blackjack game and the result from running our mechanism using either the Q-values or the action as the metric. In each sub-graph, the top figure shows the state, and the usable ace is highlighted in red if present. The bottom figure shows the importance of each feature. The solid bars are the Q-value-based importance and the hatched bars are the action-based importance. Note that at step 1, the importance for the previous hand, previous dealer, previous ace, and previous action are not applicable since there is no previous state for the first state.}
  \label{gp:blackjack}
\end{figure*}

\paragraph{Using Q-values as Metric}
The solid bars in Fig.~\ref{gp:blackjack} on the next page show the result of Q-value-based importance based on Eq.~(\ref{eq:impor_q}).  We interpret the result as follows: (1) The importance of all features are highest at step 1. This is because state 1 is closest to the decision boundary of the policy, and thus applying a perturbation at this step is easier to change the Q-value distribution; (2) The importance of \texttt{dealer} and \texttt{dealer\_prev} are the same throughout the trajectory. This is due to the fact that \texttt{dealer} and \texttt{dealer\_prev} are always the same. Thus, applying a perturbation on \texttt{dealer\_prev} will have the same effect as applying a perturbation on \texttt{dealer} assuming changing \texttt{dealer\_prev} won't incur a change in the previous action; (3) A similar phenomenon can be observed between \texttt{hand} and \texttt{hand\_prev}. Increasing the hand at step $t-1$ by one will have the same outcome as drawing a card with one higher value at $t$. The occasional difference comes from the change in \texttt{hand\_prev} causing \texttt{a\_prev} to change; (4) The importance of \texttt{ace} is highest at steps 2 and 5. In both of these two states, changing if the player has an ace or not while keeping other features the same will change the best action and a larger difference in the Q-values, which causes the importance to be higher.

\paragraph{Using Action as Metric}
The hatched bars in Fig.~\ref{gp:blackjack}  show the result of action-based importance based on Eq.~(\ref{eq:impor_ve}).
The importance is more ``bursty", and features, such as \texttt{hand}, have an importance of zero in the majority of the steps since a perturbation of size one could not trigger a change in the action. However, intuitively, \texttt{hand} is crucial to the agent's decision-making. Therefore, in this case, we note that the Q-value-based method produces a more reasonable explanation in this example.

\paragraph{Multi-Step Temporal Importance}
We cascade the causal graph of blackjack in Fig. \ref{gp:blackjack_scm}  to estimate the impact of the past states and actions on the current action, and the full SCM is shown in Fig.~\ref{gp:blackjack_concat_scm} in Appendix \ref{app:black}.
Fig.~\ref{gp:blackjack_multistep} shows the results of Q-value-based importance. The importance of $A_4$ on itself is omitted since it will always be one regardless of any other part of the graph. We interpret the results as follows: (1) The importance of \texttt{hand}$_\tau$ and \texttt{dealer}$_\tau$ is flat over time. As discussed above, perturbing these two features at any given step will mostly change the last state in the same way, resulting in constant importance; (2) The importance of the action $a_\tau$ increases as $\tau$ gets closer to the last step $t=4$. An action taken far in the past should generally have a smaller impact on the current action, which corresponds to the increasing importance for \texttt{a}$_\tau$ in our explanation.

\subsection{Additional Evaluation}

We also evaluate our scheme in a more complex RL environment, Lunar Lander, in Appendix \ref{app:lunar_lander}. Lunar Lander is a simulation testing environment developed by OpenAI Gym~\cite{brockman2016openai}. The simulation shows that our scheme can explain some specific phases(state) of the spaceship in the landing process.

\begin{figure}[htbp]
  \centering
  \includegraphics[width=0.7\linewidth]{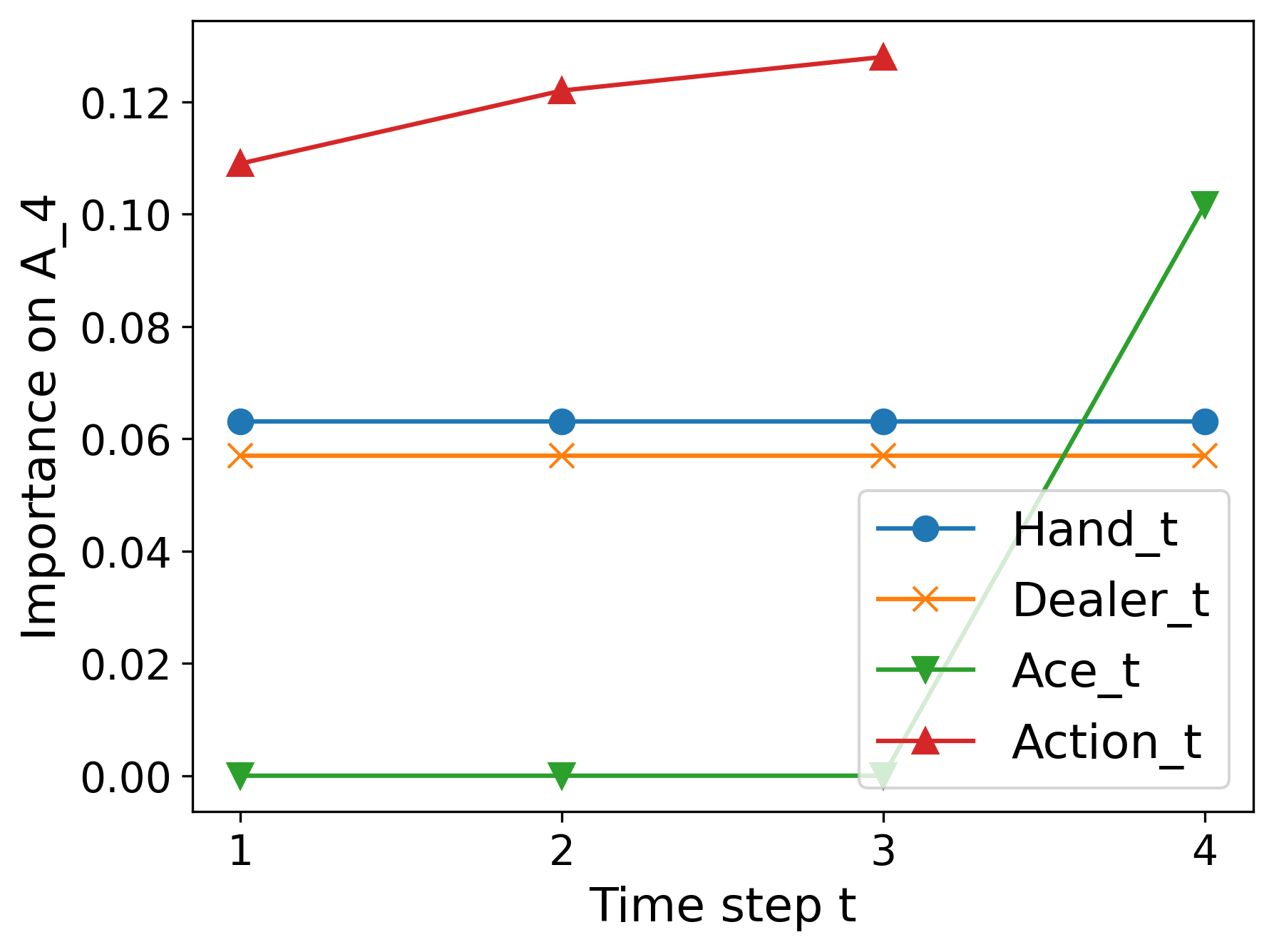}
  \caption{The Q-value-based temporal importance on $A_4$ for all state features and actions at past time steps in the Blackjack experiment.}
  \label{gp:blackjack_multistep}
\end{figure}

\section{Discussions}

Our causal importance explanation mechanism is a post-hoc explanation method that uses  data collected by an already learned policy. We focus on providing local explanations based on a particular state and action. Counterfactual reasoning is required to recover the exogenous variables and estimate the effect on the given state and action. In this case, the intervention operation is not enough to achieve this goal, as it can only evaluate the average results (population) over the exogenous variables, which is not a local explanation for the given state.

\paragraph{Intra-state Relations}

One crucial characteristic of our method is that we consider intra-state relations when computing the importance, which is essential in accurately quantifying the impact of a state feature on the action. Although the MDP defines that a state feature at a certain time step cannot affect another state feature at the same step, it is essential to consider  causal relationships within state features when measuring their impact if we use causal intervention or associational perturbation. Since these types of methods require modifying the value of a specific  state feature, it should subsequently affect the value of other state features based on real-world causality. For instance, in the collision avoidance problem (Section \ref{sec:bang_bang}), the distance to the end ($D_t$) will change in response to the distance from the start ($X_t$), and in the crop irrigation problem (Section \ref{sec:crop_simulation}), the crop weight ($C_t$) will vary based on the humidity level ($H_t$). Ignoring the intra-state causality can lead to an invalid state after the intervention, resulting in inaccurate importance estimates for the given state feature. Hence, we formulate the intra-state relations in the SCM to provide more accurate and comprehensive explanations of the problem.

\paragraph{Additive Noise Assumption}

With the additive noise assumption in Eq.~(\ref{eq:additive}), the exogenous variable (noise) can be fully recovered and used for counterfactual reasoning. We note that the full recovery noise assumption can be relaxed for our mechanism. In the case where the exogenous variables have multiple values (not deterministic), we can generalize our definition of importance vector in Eq.~(\ref{eq:impor_ve}) by replacing the first term with the expectation over different values of exogenous variables  using probabilistic counterfactual reasoning \cite{glymour2016causal}. Furthermore, the additive noise assumption is not mandatory. We can use bidirectional conditional GAN  \cite{jaiswal2018bidirectional} to model the structure function and use its noise to conduct counterfactual reasoning and obtain the importance vector.

\paragraph{Known SCM Skeleton Assumption}

Our approach is based on the assumption that the SCM skeleton is known, which can be obtained either through background knowledge of the problem or learned using causal discovery algorithms. Causal discovery aims to identify causal relations by analyzing the statistical properties of purely observational data. There are several causal discovery algorithms available, including the classical constraint-based PC algorithm ~\cite{spirtes2000causation}, algorithms based on linear non-Gaussian models ~\cite{shimizu2006linear}, and algorithms that use the additive noise assumption ~\cite{hoyer2008nonlinear,peters2014causal}. These algorithms can be used to learn the SCM skeleton from observational data, which can then be used in our method to quantify the impact of state features and actions on the outcome. There are also existing toolboxes such as \cite{kalainathan2019causal} and \cite{zhang2021gcastle} that can be easily applied directly to data to identify the SCM structure.

\paragraph{Perturbation}
In addition to the method we employed in the simulation, which averages the importance derived from both positive and negative $\delta$, maximizing them is also a viable option.To compute the causal importance vector defined in Eq.~(\ref{eq:impor_ve}), we need to choose a perturbation value  $\delta$. As shown in Table~\ref{tb:score}, the importance may depend on $\delta$. Therefore, it is not meaningful to compare  importance vectors calculated with different $\delta$. This is a common issue of perturbation-based algorithms, including the saliency map method. In our case, $\delta$ should be as small as possible but still be computationally feasible. More detailed sensitivity analysis and  normalization on the perturbation value $\delta$ can be found in Appendix~\ref{sec:sens}.

\paragraph{Limitations}

Our study has limitations when the state space has high dimensions, for example, in visual RL, where state features are represented as images. Image data is inherently high-dimensional, with multiple features that can interact in complex ways. The SCMs we used may struggle to fully capture the complexity of these interactions, especially when a large number of variables are involved. To address this issue, we suggest utilizing the algorithm of causal discovery in images \cite{lopez2017discovering} and representation learning \cite{yang2021causalvae}. Further work is needed to explore this direction.

Another question that might be raised is what will happen if the trained SCM is not perfect. An imperfect SCM will cause the counterfactual reasoning result to be biased, and thus affecting the final importance. One potential solution is quantifying the uncertainty of the explanation. If the explainer can output its confidence on top of the importance score, users can identify potential out-of-distribution samples where our explanation framework might fail. To achieve this, we need to separate aleatoric uncertainty (which comes from the inherent variability in the environment) and epistemic uncertainty (which represents the imperfection of the model) \cite{gawlikowski2021survey}. Our use of SCM may help us to differentiate the two, and this is one of the directions we are currently exploring.

\section{Conclusion}
In this paper, we have developed a causal explanation mechanism that quantifies the causal importance of states on actions and their temporal importance. Our quantitative and qualitative comparisons show that our explanation can capture important factors that affect actions and their temporal importance. This is the first step towards causally explaining RL policies. In future work, it will be necessary to explore different mechanisms to quantify causal importance, relax existing assumptions, build benchmarks, develop human evaluations, and use the explanation to improve evaluation and RL policy training. 

\newpage
\bibliography{sn-bibliography}

\newpage
\appendix
\onecolumn

\renewcommand{\thefigure}{\arabic{figure}}
\setcounter{figure}{9}

\section{Additional Experiments and Details}
\label{app:addition_experiments_and_details}

In this section, we provide additional details regarding the crop irrigation problem, the collision avoidance problem, and the Blackjack experiments. Furthermore, we describe our results on an additional testing environment, Lunar Lander.

All experiments were conducted on a machine with 8 NVIDIA RTX A5000 GPU, an dual AMD EPYC 7662 CPU, and 256 GB RAM.

\subsection{Crop Irrigation}
\label{app:crop}

This section contains details of the crop irrigation experiment.
\paragraph{System dynamics}
\begin{align*}
    \textrm{Precipitation} &= U(0,1)\\
   	\textrm{SolarRadiation} &= U(0,1)\\
    \textrm{Humidity} &= 0.3 \cdot \textrm{Humidity}_\textrm{prev} + 0.7 \cdot \textrm{Precipitation}\\
    \textrm{CropWeight} &= \textrm{CropWeight}_\textrm{prev} \\
    & + 0.07 \cdot \big(1-(0.4 \cdot \textrm{Humidity} + 0.6 \cdot \textrm{Irrigation} - \textrm{Radiation}^2)^2\big) \\
    & + 0.03 \cdot U(0,1)
\end{align*}
The change in \textrm{CropWeight} at each step is determined by humidity, irrigation and radiation, and maximum growth is achieved when $0.4 \cdot \textrm{Humidity} + 0.6 \cdot \textrm{Irrigation} = \textrm{Radiation}^2$. An additional exogenous variable is also included in the change of \textrm{CropWeight}. This can be regarded as some unobserved confounders that affect the growth that are not included in the system dynamics, such as \texttt{CO}$_2$\texttt{Concentration} or the temperature.

\paragraph{Policy}
\begin{equation*}
    \textrm{Irrigation} = (\textrm{Radiation}^2 - 0.4 \cdot \textrm{Humidity}) \cdot (1.6 \cdot \textrm{CropWeight} + 0.2) / 0.6
\end{equation*}
The policy we used is a suboptimal policy that multiplies an additional coefficient $1.6 \cdot \textrm{CropWeight} + 0.2$ on the optimal policy. This will cause the irrigation value to be less than optimal when \textrm{CropWeight} is less than 0.5, and   more than optimal and vice versa.

\paragraph{Training}
We use a neural network to learn the causal functions in the SCM. The network has three fully-connected layers, each with a hidden size of four. We use Adam with a learning rate of $3 \times 10^{-5}$ as the optimizer. The training dataset consists of 1000 trajectories (10000 samples) and the network is trained for 50 epochs. 

\paragraph{Perturbation}
The perturbation value $\delta$ used in the intervention is $0.1$ w.r.t. the range of each value.

\begin{figure}[htp]
\centering
\captionsetup{justification=centering}
\begin{subfigure}[t]{0.32\linewidth}
  \includegraphics[width=\linewidth]{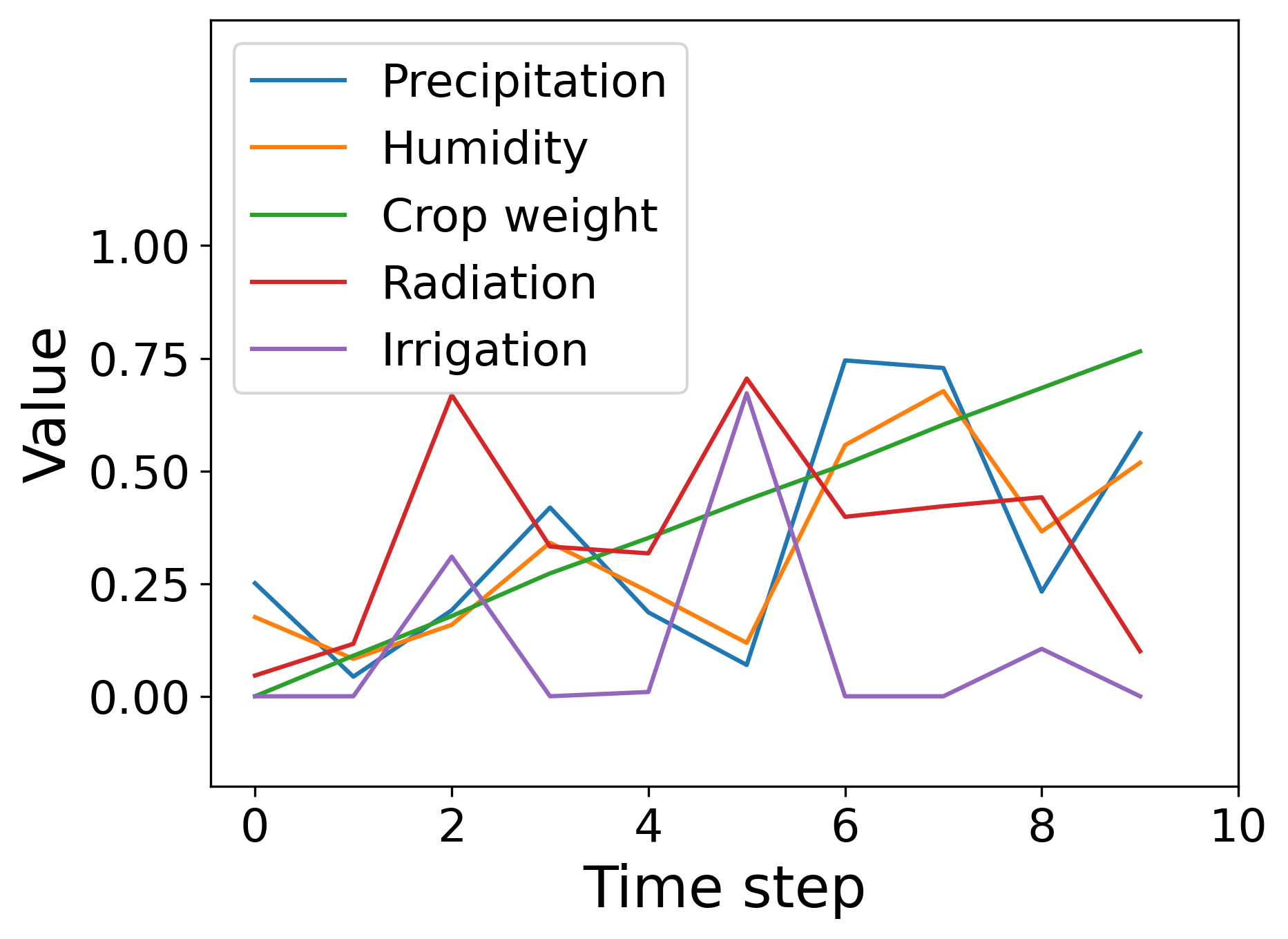}
  \caption{A trajectory of the crop irrigation problem}
  \label{gp:crop_traj}
\end{subfigure}
\begin{subfigure}[t]{0.32\linewidth}
  \includegraphics[width=\linewidth]{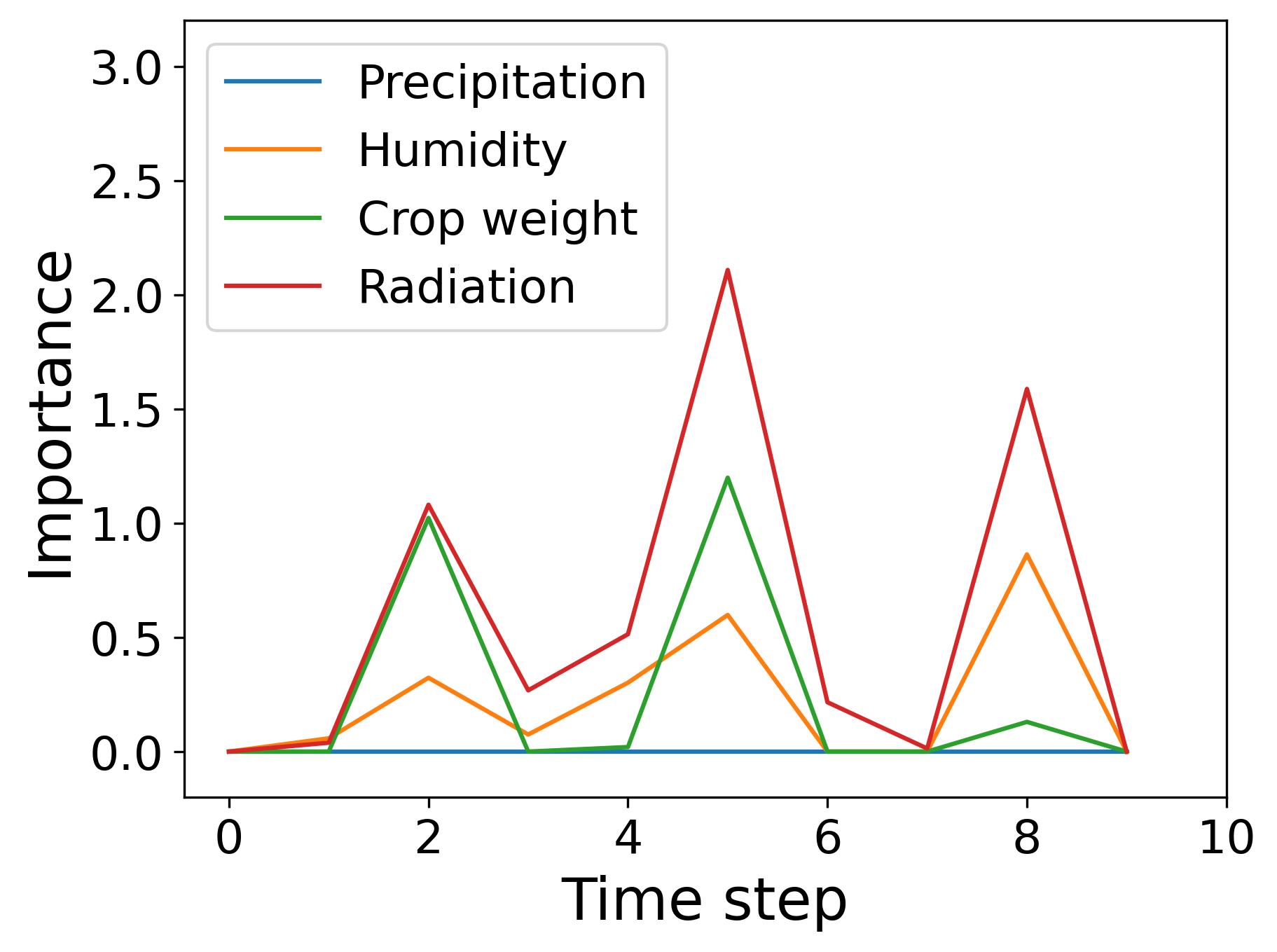}
  \caption{Saliency map method result}
  \label{gp:crop_saliency}
\end{subfigure}
\begin{subfigure}[t]{0.32\linewidth}
  \includegraphics[width=\linewidth]{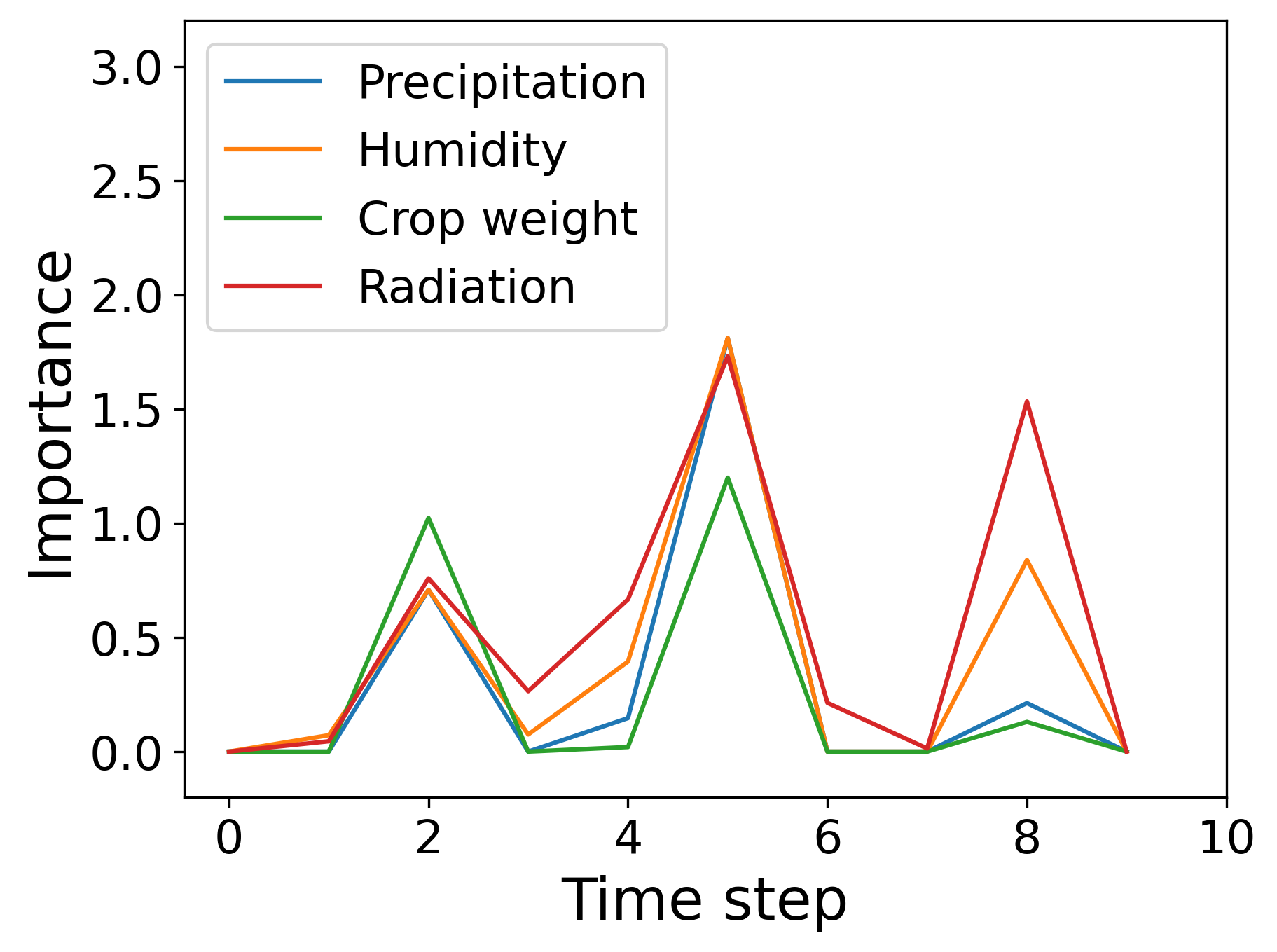}
  \caption{Causal-based importance result}
  \label{gp:crop_causal}
\end{subfigure}
\caption{Importance vector for state in crop irrigation problem}
\label{gp:grop_compare_time}
\end{figure}


\subsection{Blackjack}
\label{app:black}
This section contains details and additional figures for the blackjack simulation.

\paragraph{System dynamics}
This simulation is done in the blackjack environment in OpenAI Gym~\cite{brockman2016openai}. The goal is to draw cards such that the sum is close to 21 but never exceeds it. Jack, queen and king have a value of 10, and an ace can be either a 1 or an 11, and an ace is called ``usable" when it can be used at an 11 without exceeding 21. We assume the deck is infinite, or equivalently each card is drawn with replacement.

In each game, the dealer starts with a shown card and a face-down card, while the player starts with two shown cards. The game ends if the player's hand exceeds 21, at which the player loses, or if the player chooses to stick, the dealer will reveal the face-down card and draw cards until his sum is 17 or higher. The player wins if the player's sum is closer to 21 or the dealer goes bust.

\begin{figure}[htp]
\centering
\includegraphics[width=.6\linewidth]{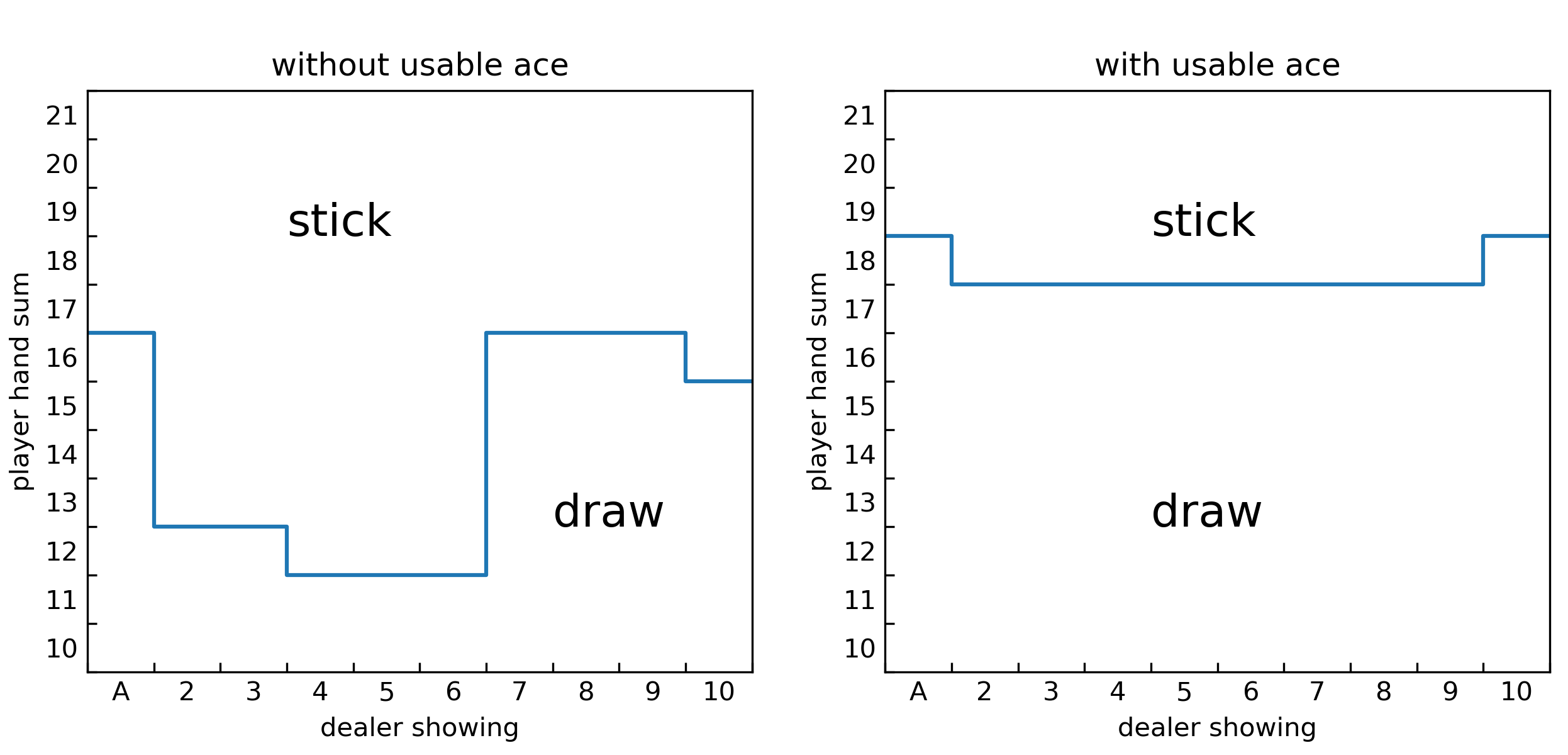}
\caption{The policy we use for the blackjack game. The blue line shows the decision boundary.}
\label{gp:blackjack_policy}
\end{figure}

\paragraph{Policy}
We trained the agent using on-policy Monte-Carlo control. Fig.~\ref{gp:blackjack_policy} shows the policy and the decision boundary.

\begin{figure}[htp]
\centering
\includegraphics[width=0.9\linewidth]{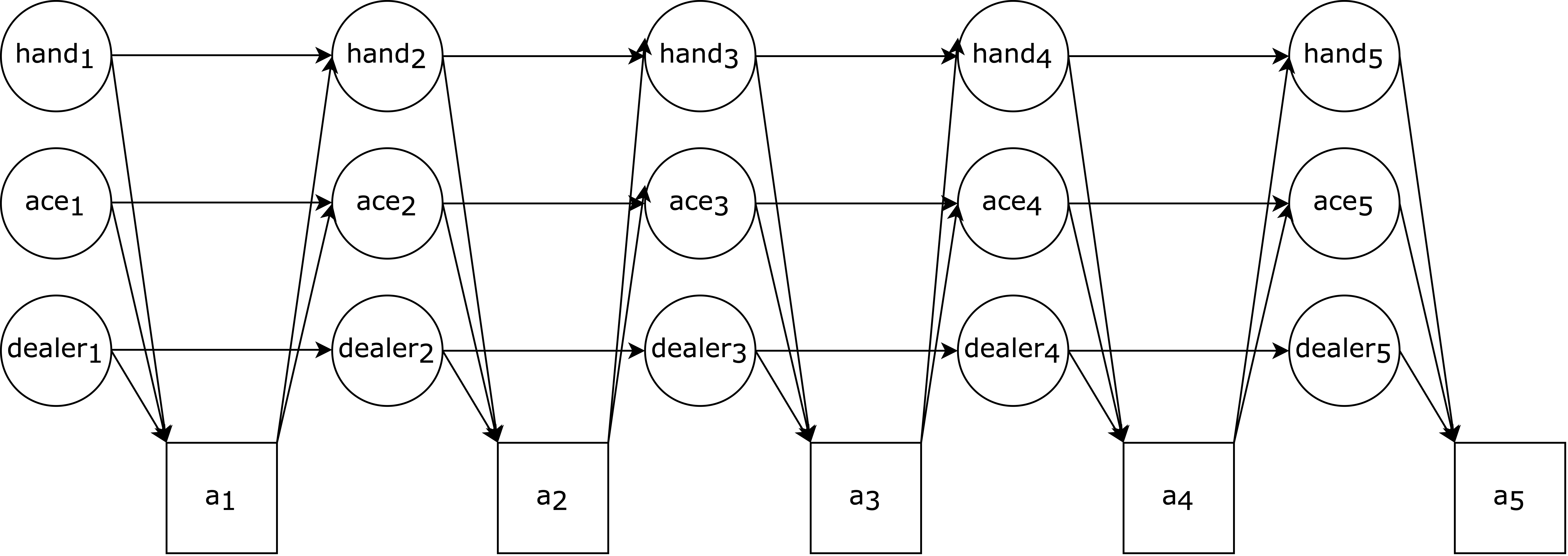}
\caption{The skeleton of the cascading SCM for a 5-step blackjack game.}
\label{gp:blackjack_concat_scm}
\end{figure}

\paragraph{SCM structure}
We assume the blackjack game has a causal structure as shown in Fig.~\ref{gp:blackjack_scm}. Additionally, Fig.~\ref{gp:blackjack_concat_scm} shows the 5-step cascading SCM we used to test the temporal importance.

\paragraph{Training}
We use a neural network to learn the causal functions in the SCM. The network has three fully-connected layers and each layer has a hidden size of four. We use Adam with a learning rate of $3 \times 10^{-5}$ as the optimizer. The training dataset consists of 50000 trajectories ($\sim$76000 samples) and the network is trained for 50 epochs.

\paragraph{Perturbation}
Since blackjack has a discrete state space, for numerical features ``hand" and ``dealer", we use a perturbation value $\delta = 1$. For the boolean feature ``ace", we flip its value as the perturbation.

\subsection{Collision Avoidance Problem}
\label{app:bang_bang}
We use the collision avoidance problem to further illustrate that our causal method can find a more meaningful importance vector than saliency map, i.e., which state feature is more impactful to decision-making.

\paragraph{System dynamics}

The state $\mathbf{S}_t$ includes the distance from the start $X_t$, the distance to the end $D_t$, and the velocity $V_t$ of the car, i.e., $\mathbf{S}_t := [V_t, X_t, D_t]$, where $V_t \leq v_{\max}$ and $v_{\max}$ is the maximum speed of the car. The action $A_t$ is the car's acceleration, which is bounded  $\vert A_t \vert \leq e_{\max}$. The state transition is defined as follows: 
\begin{align*}
V_{t+1} &:= V_t + A_t\Delta t\\
X_{t+1} &:= X_t + V_t\Delta t + \frac{1}{2}A_t\Delta t^2\\
D_{t+1} &:= X_\text{goal} - X_{t+1}
\end{align*}
The objective of the RL problem is to find a policy $\pi$ to minimize the traveling time under the condition that the final velocity is zero at the endpoint (collision avoidance).

\paragraph{Policy}
An RL agent learns the following {\bf optimal} control policy also known as the bang-bang control  (optimal under certain technical conditions) defined as Eq. (\ref{eq:bangpolicy})

\paragraph{SCM structure}
We use Fig.~\ref{gp:bang_caual} as the SCM skeleton and use linear regression to learn the structural equations as the entire dynamics are linear. 

\paragraph{Perturbation}
The perturbation value $\delta$ used in the intervention is $0.1$ after normalization.

\subsection{Lunar Lander}
\label{app:lunar_lander}

\paragraph{System dynamics}
Lunar lander problem is a simulation testing environment developed by OpenAI Gym~\cite{brockman2016openai}. The goal is to control a rocket to land on the pad at the center of the surface while conserving fuel. The state space is an 8-dimensional vector containing the horizontal and vertical coordinates, the horizontal and vertical speed, the angle, the angular speed, and if the left/right leg has contacted or not. 

The four possible actions are to fire one of its three engines: the main, the left, or the right engine, or to do nothing. 

The landing pad location is always at $(0, 0)$. The rocket always starts upright at the same height and position but has a random initial acceleration. The shape of the ground is also randomly generated, but the area around the landing pad is guaranteed to be flat.

\paragraph{Policy}
We train our RL policy using DQN~\cite{van2016deep}. 

\begin{figure}[htp]
\centering
\includegraphics[width=.7\linewidth]{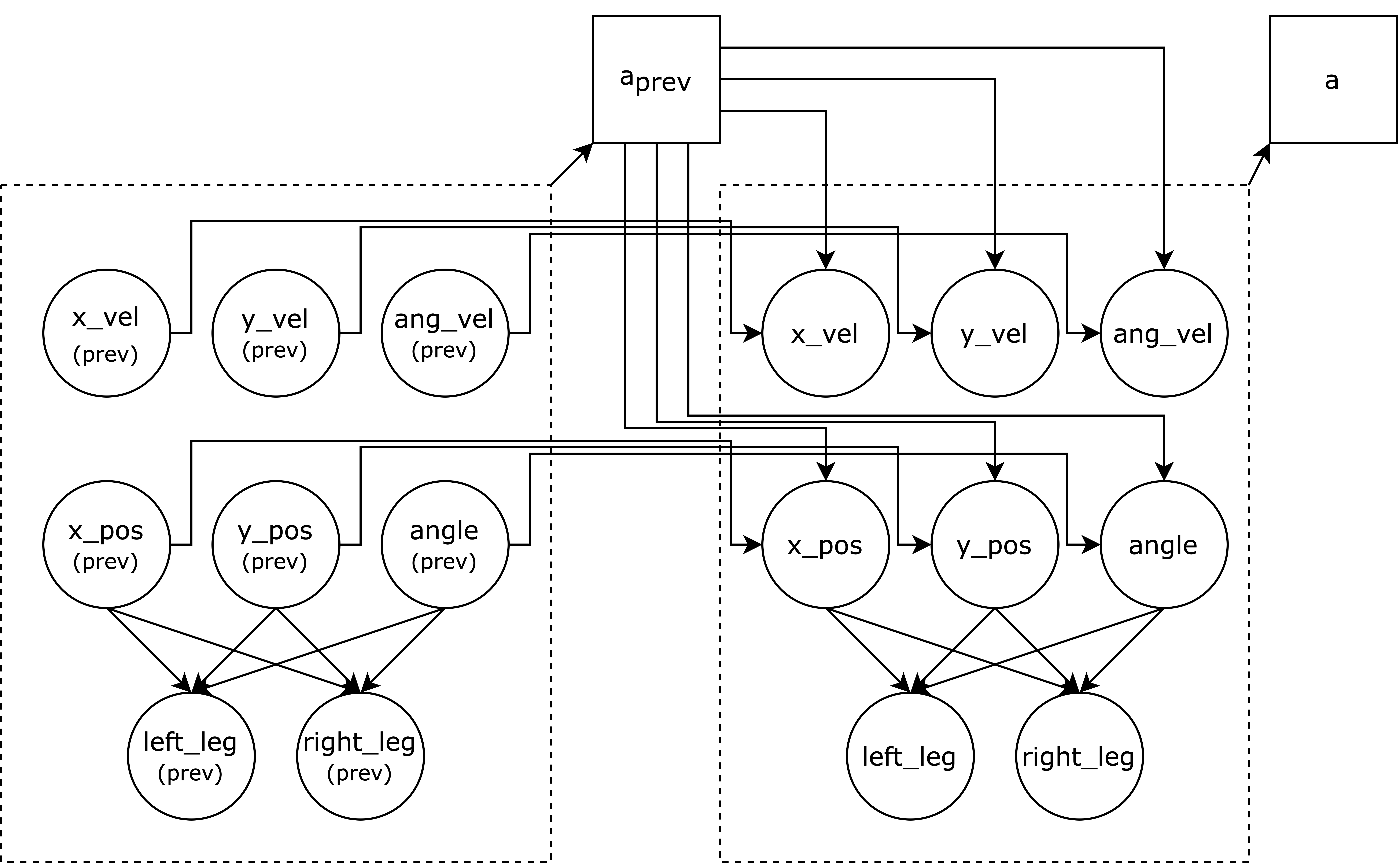}
\caption{The causal structure of lunar lander that includes previous state and actions. There should also be edges from each feature to the action at its time step, e.g. edges from \texttt{x\_pos\_prev} to \texttt{a\_prev}, or from \texttt{x\_pos} to \texttt{a}. These edges are not shown in this graph for simplicity.}
\label{gp:lunar scm}
\end{figure}

\paragraph{SCM structure}
We use the Fig.~\ref{gp:lunar scm} as the skeleton of SCM. The structural functions are learned with linear regression using 100 trajectories ($\sim$25000 samples).

\begin{figure}[htp]
\centering
\begin{subfigure}[t]{.3\linewidth}
    \centering
    \includegraphics[width=\linewidth]{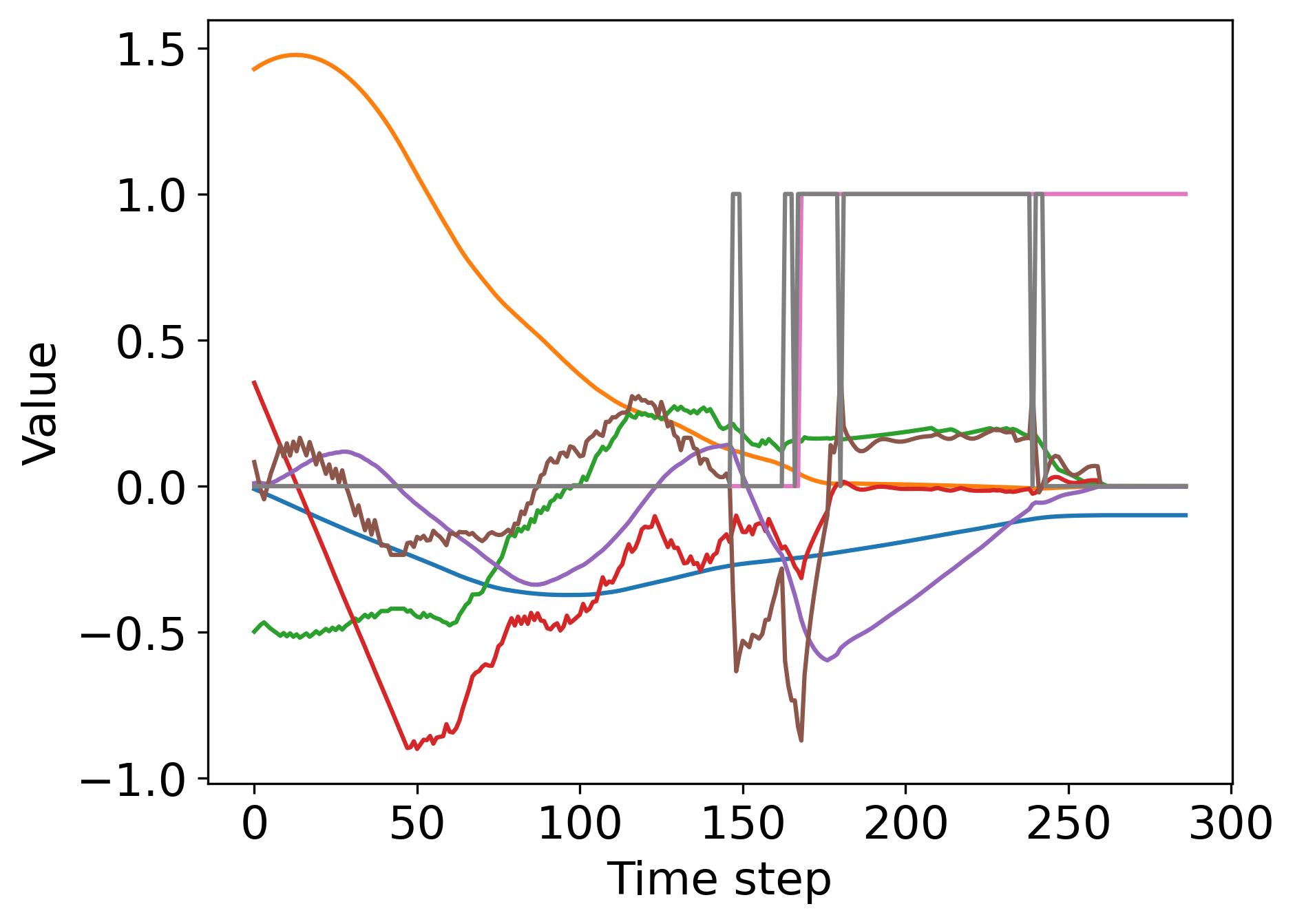}
    \caption{The lunar lander instance.}
    \label{gp:lunar_lander_trajectory}
\end{subfigure}\quad
\begin{subfigure}[t]{.3\linewidth}
    \centering
    \includegraphics[width=\linewidth]{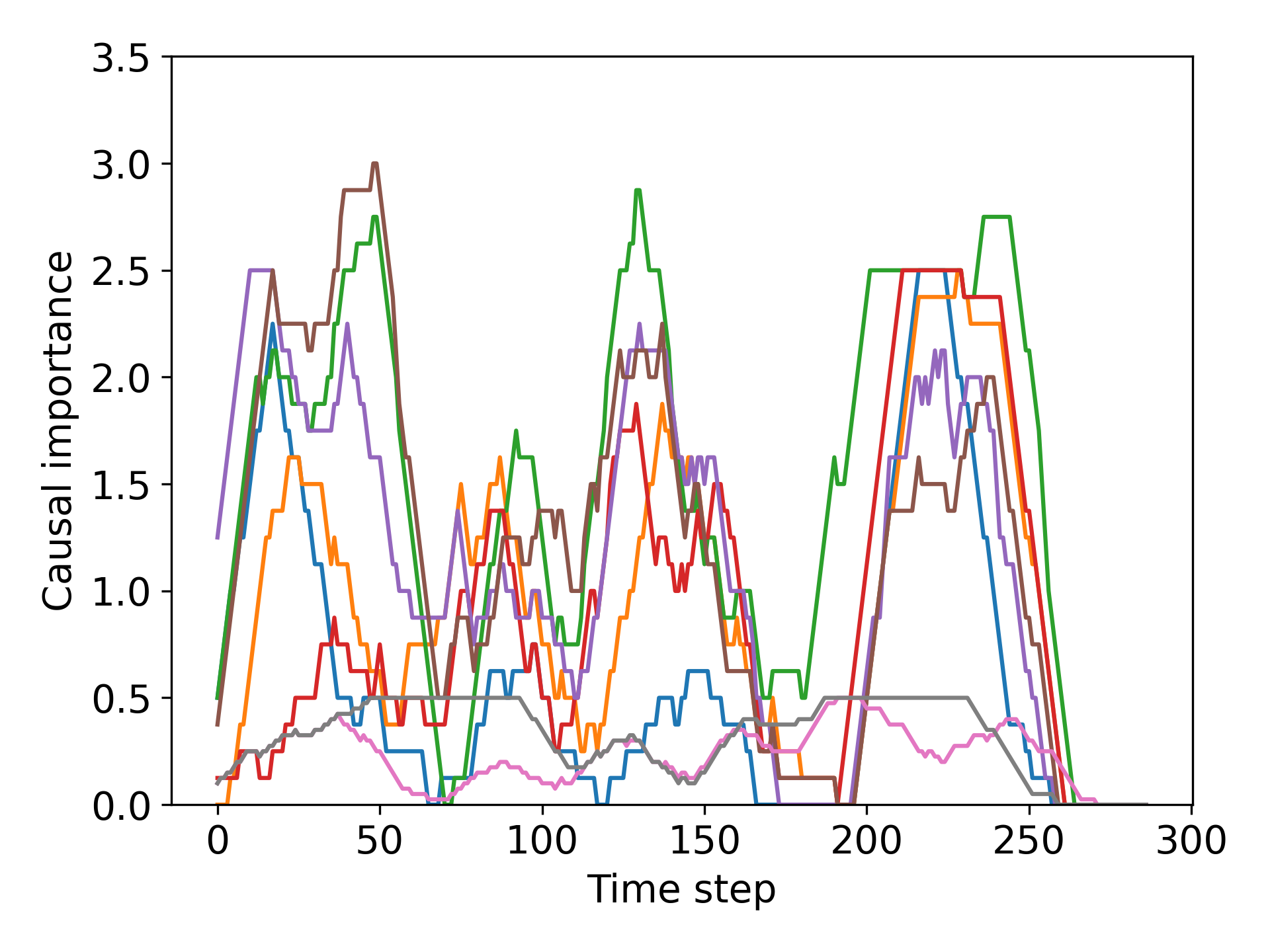}
    \caption{Causal importance vector the for the current-step features on the trajectory in Fig.~\ref{gp:lunar_lander_trajectory}.}
    \label{gp:lunar_lander_curr}
\end{subfigure}\quad
\begin{subfigure}[t]{.3\linewidth}
    \centering
    \includegraphics[width=\linewidth]{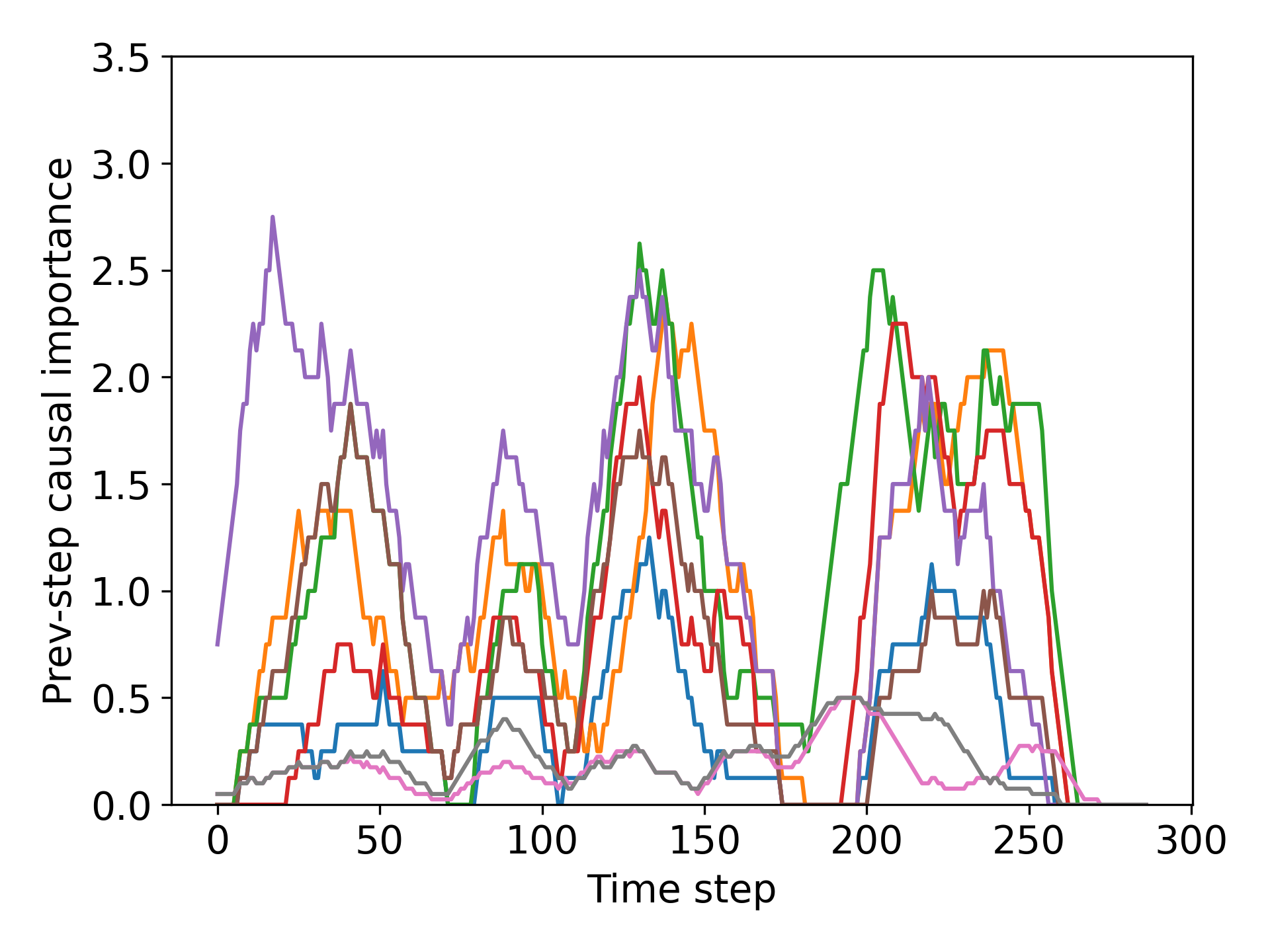}
    \caption{Causal importance vector the for the previous-step features on the trajectory in Fig.~\ref{gp:lunar_lander_trajectory}.}
    \label{gp:lunar_lander_prev}
\end{subfigure}
\begin{subfigure}{.75\linewidth}
    \centering
    \includegraphics[width=\linewidth]{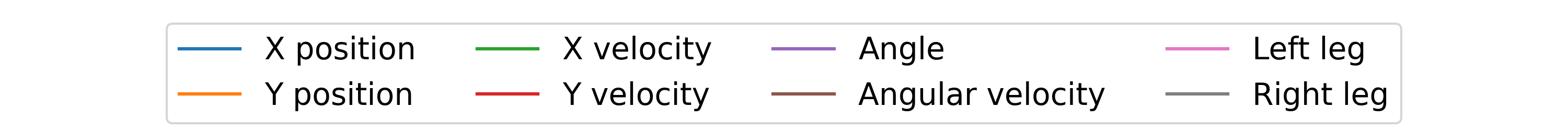}
\end{subfigure}
\caption{A lunar lander trajectory instance we used to evaluate our algorithm and the corresponding causal importance vector. The ``freefall phase" is roughly between steps 0-70, ``adjusting phase" is between steps 70-170, and ``touchdown phase" is from about step 170 to the end.}
\label{gp:lunar_lander}

\end{figure}

\begin{figure}[htp]
\captionsetup{justification=centering}
\centering
  \centering
  \begin{subfigure}[t]{.3\linewidth}
    \centering
    \includegraphics[width=\linewidth]{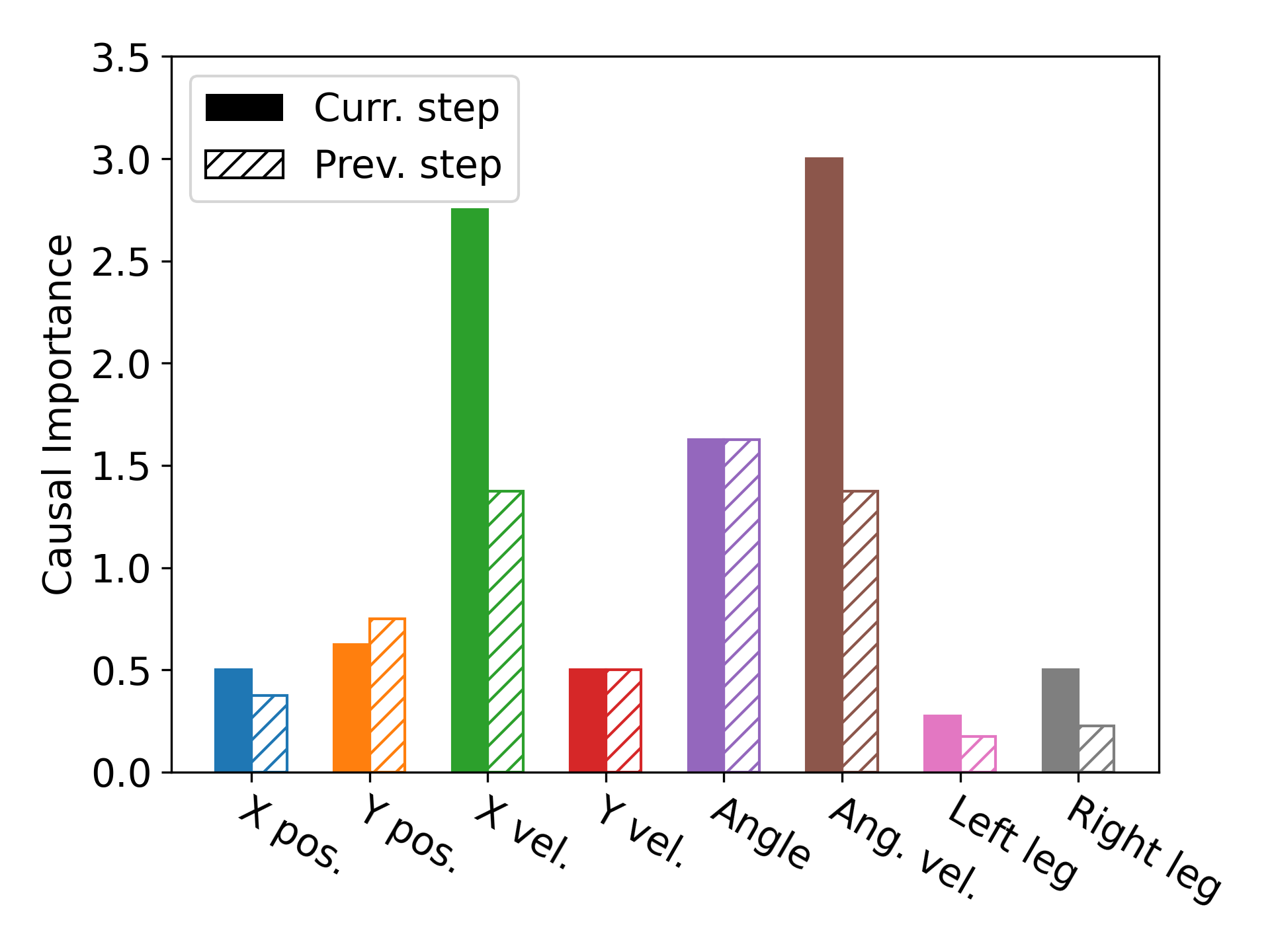}
    \caption{Importance vector during the ``free fall phase" (step 48).}
    \label{gp:lunar_lander_48}
  \end{subfigure}\quad
  \begin{subfigure}[t]{.3\linewidth}
    \centering
    \includegraphics[width=\linewidth]{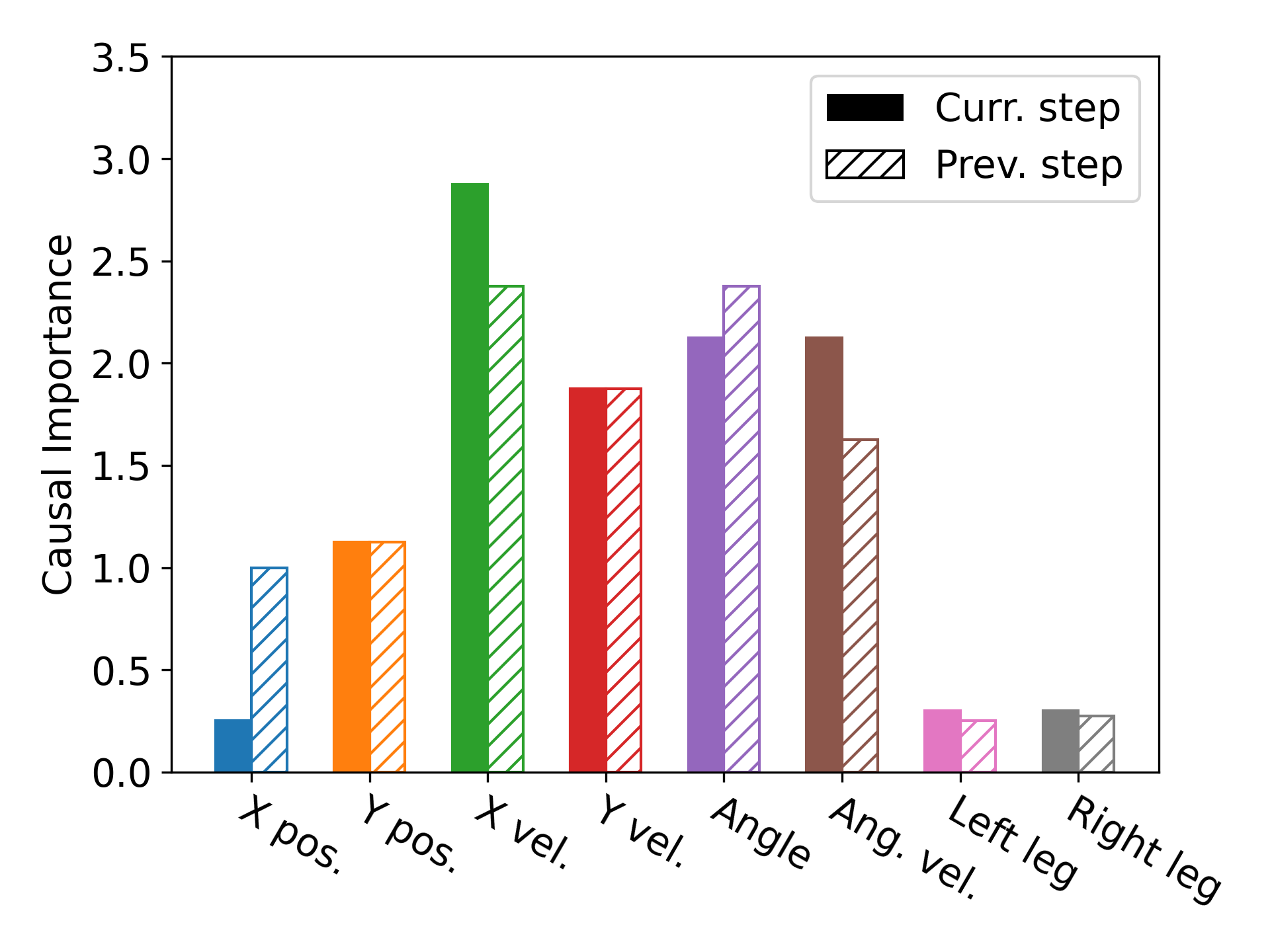}
    \caption{Importance vector during the ``adjusting phase" (step 129).}
    \label{gp:lunar_lander_129}
  \end{subfigure}\quad
  \begin{subfigure}[t]{.3\linewidth}
    \centering
    \includegraphics[width=\linewidth]{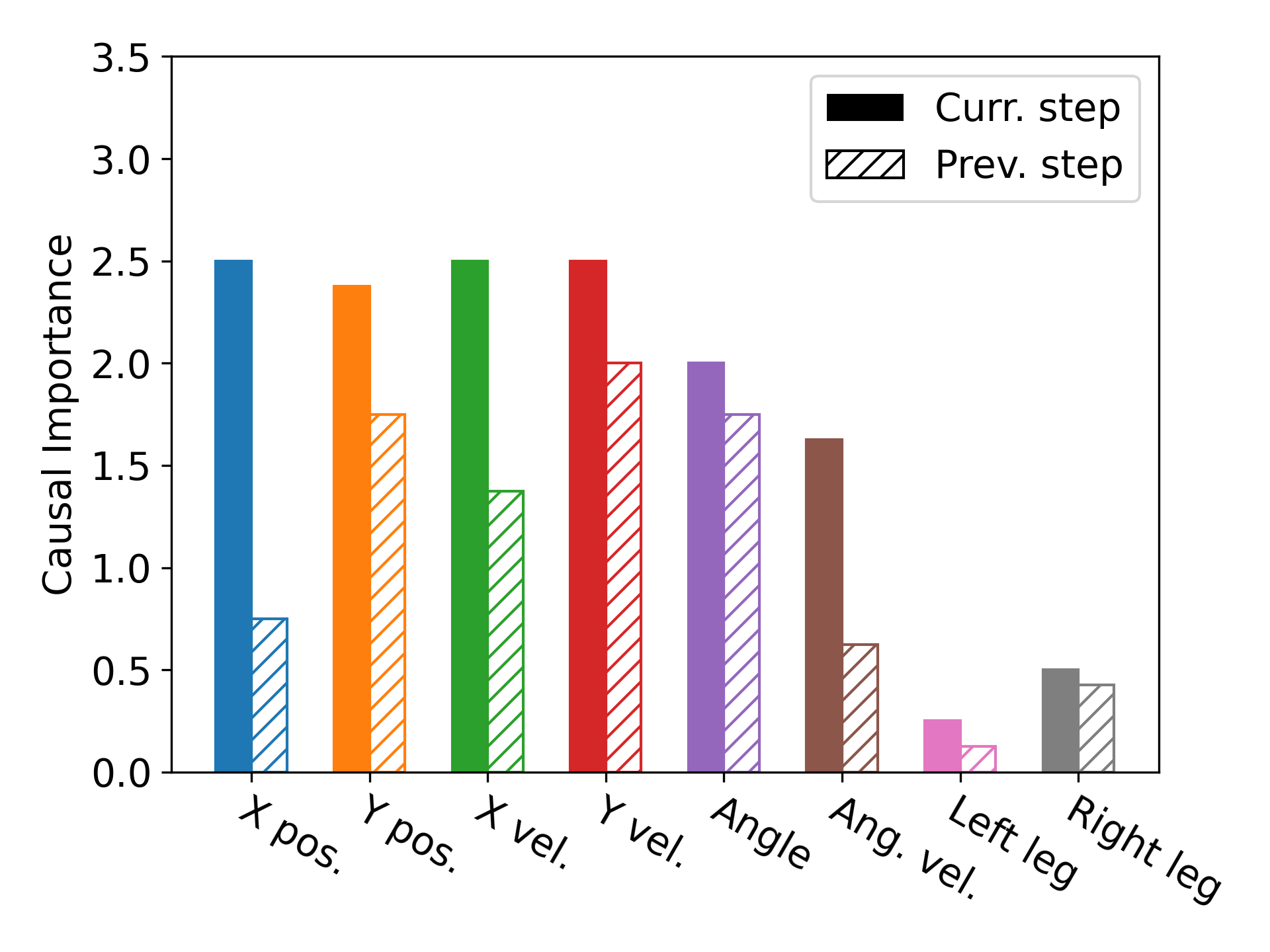}
    \caption{Importance vector during the ``touchdown phase" (step 216).}
    \label{gp:lunar_lander_216}
  \end{subfigure}
\caption{The importance vector on lunar lander calculated using our method and a comparison with the saliency map method. The solid bars in the first three figures representing the importance of the current-step features and the shaded bars are for the previous-step features.}
\label{gp:lunar_lander_points}
\end{figure}

\paragraph{Evaluation}
Fig.~\ref{gp:lunar_lander} shows a trajectory of the agent interacting with the lunar lander environment and the corresponding causal importance using our mechanism. We notice that our mechanism discovers three importance peaks, and we explain this as  the agent's decision-making during the landing process consisting of three phases: a ``free fall phase'', in which the agent mainly falls straight and slightly adjusts its angle to negate the initial momentum; an ``adjusting phase", in which the agent mostly fires the main engine to reduce the Y-velocity; and a ``touchdown phase", during which the lander is touching the ground and the agent is performing final adjustments to stabilize its angle and speed. Fig.~\ref{gp:lunar_lander_48},~\ref{gp:lunar_lander_129} and~\ref{gp:lunar_lander_216} show our causal importance vector during each of the three phases. We notice that during the ``free fall phase", features such as angle, angular velocity and x-velocity are more important since the agent needs to rotate to negate the initial x-velocity. However, as the rocket approaches the ground during the ``adjusting phase", we find an increase in importance for y-velocity since a high vertical velocity is more dangerous to control when the rocket is closer to the ground. In the last ``touchdown phase", a large x-position and x-velocity importance can be observed as a change in those features is highly likely to cause the lander to fail to land inside the designated landing zone. Since the lander is already touching the ground, it will take much more effort for the agent to adjust compared to when the lander is still high in the air. 

\begin{figure}[htp]
    \centering
    \includegraphics[width=.4\linewidth]{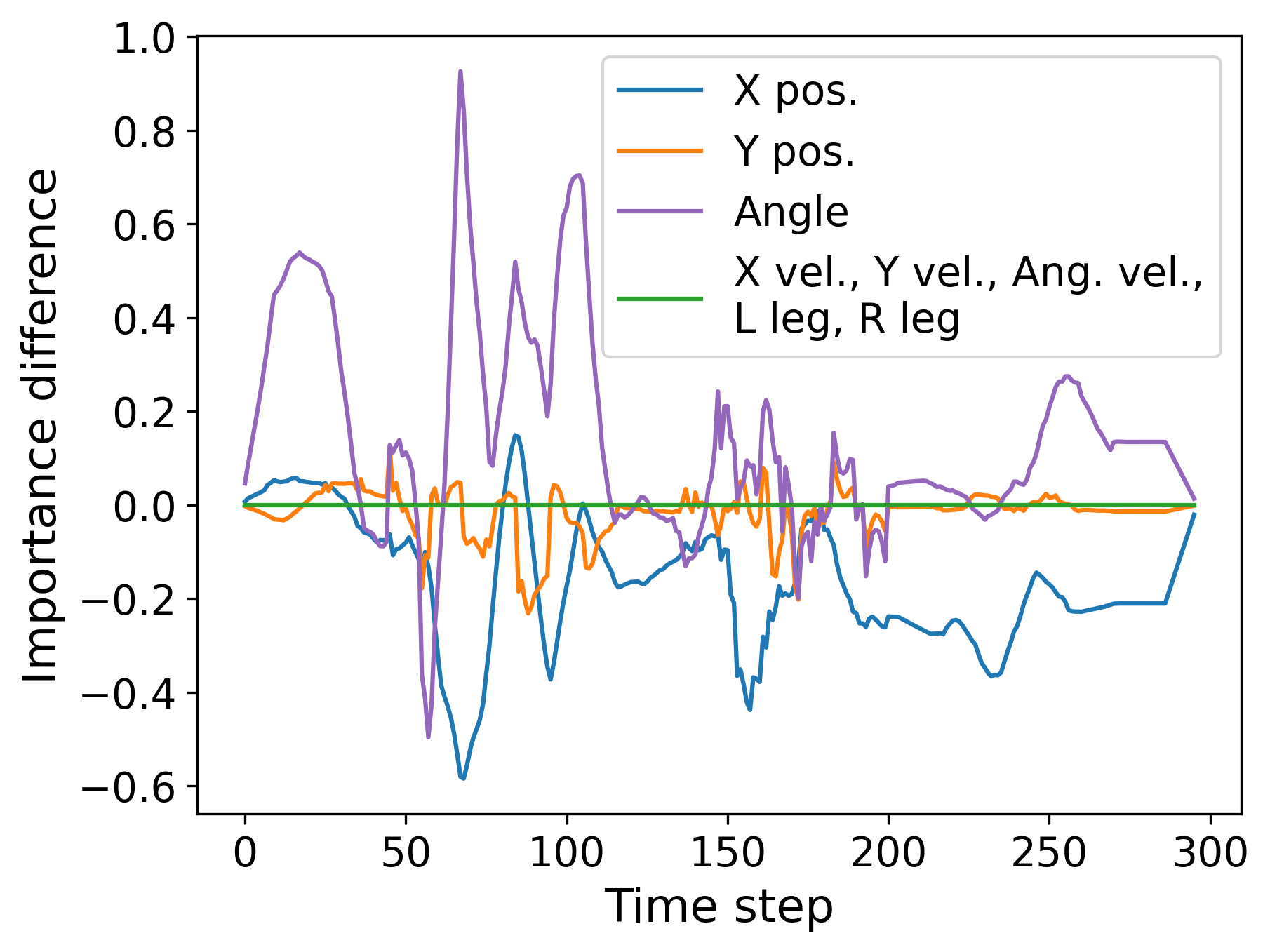}
    \caption{Difference between our method and the saliency map method for current-step features.}
    \label{gp:lunar_lander_diff}
  \end{figure}

The results are similar to those of saliency-based algorithms~\cite{greydanus2018visualizing}, and Fig.~\ref{gp:lunar_lander_diff} shows the difference in importance vector between our algorithm and saliency-based algorithm. Note that differences only occur for the positions and the angle. This is because other features don't have any additional causal paths to the action besides the direct connection. Therefore, the intervention operation is equivalent to the conditioning operation for these features. The features position and angle have an additional causal path through the legs, which causes the difference. Notably, our method captures higher importance for angle, which we interpret as that the landing angle is crucial and is actively managed by the agent. 

We are also able to compute the importance of the features in the previous steps, and Fig.~\ref{gp:lunar_lander_prev} and the shaded bars in Fig.~\ref{gp:lunar_lander_points} represent such importance vectors. The previous-step importances are rather similar to those of the current-step features since the size of the time step is comparatively small. However, our algorithm captures that during the ``adjusting phase", the previous-step importance for the angle is in general higher than the current-step importance, as changing the previous angle may have a cascading effect on the trajectory and is especially important to the agent when it is actively adjusting the angle. 


\section{Sensitivity Analysis}
\label{sec:sens}
This section performs a sensitivity analysis on how the perturbation amount affects the result of our explanation.

For action-based importance, too small of a perturbation may not yield a meaningful result. This is due to the fact that, depending on the environment and the policy, a too small perturbation may fail to trigger a noticeable change in the action, resulting in a zero importance. This differs from the zero importance case where the policy disregards the feature when making decisions. In our experiments, we use 0.01 with respect to the range of the features for continuous features and the smallest unit for discrete features.

In general, using different perturbation amounts $\delta$ on the same state in the same SCM may result in different importance vectors, and vectors calculated using different $\delta$ cannot be meaningfully compared. However, if we desire the importance of using different $\delta$ to be more on the same level, we suggest finding the highest importance across all features and all time steps and normalizing all results by said number. Section~\ref{sec:sensitivity_bang_bang} contains an example comparing the importance score with and without the aforementioned normalization.

\begin{figure}[htp]
\centering
\includegraphics[width=0.4\linewidth]{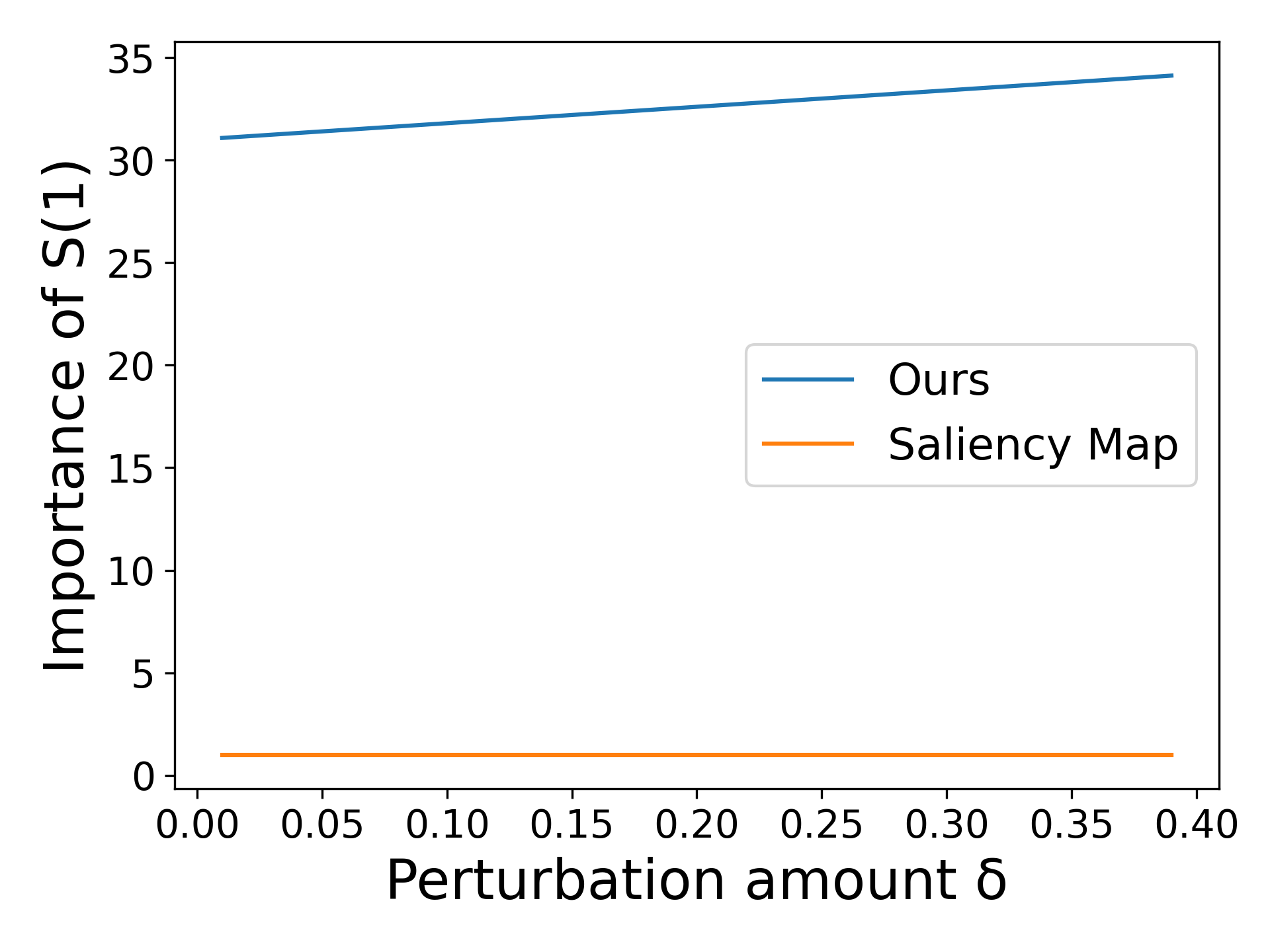}
\caption{The importance vector of $\mathbf{S}^{(1)}$ from both our method and the saliency map method with respect to the perturbation amount.}
\label{gp:one_step_sensitivity}
\end{figure}

\subsection{One-step MDP}
\label{sec:sensitivity_one_step}
As we demonstrated in the example of one-step MDP in Fig.~\ref{gp:thexp} and Table~\ref{tb:score}, our importance vector will sometimes be affected by the perturbation amount. For this experiment, we use Fig.~\ref{gp:thexp} as the skeleton and the following settings. The constants are 
\[
	c_1 = 1, c_2 = -2, c_3 = 3, c_{12} = 2, c_p = -1
\]
We use unit Gaussian distributions as the exogenous variables and the values are
\[
	u_1 = 0.50, u_2 = -0.14, u_3 = 0.65, u_p = 1.52, u_a = -0.23
\]
The state value and the corresponding action are then
\[
	\mathbf{s}^{(1)} = 0.50,
	\mathbf{s}^{(2)} = 0.86,
	\mathbf{s}^{(3)} = -0.88,
	v_p = 1.52,
	a = 3.83
\]
The result of running our method and the saliency map method on the feature $\mathbf{S}^{(1)}$ is shown in Fig.~\ref{gp:one_step_sensitivity}. Same as in Table~\ref{tb:score}. Our algorithm is linear w.r.t. $\delta$ while the saliency map result is constant. The increased importance comes from the causal link $\mathbf{S}^{(1)}\!\to\!\mathbf{S}^{(2)}\!\to\! A$, which also introduces the linear relationship.

\subsection{Collision Avoidance}
\label{sec:sensitivity_bang_bang}
\begin{figure}[htp]
\centering
\begin{subfigure}{0.45\textwidth}
\centering
	\includegraphics[width=0.75\textwidth]{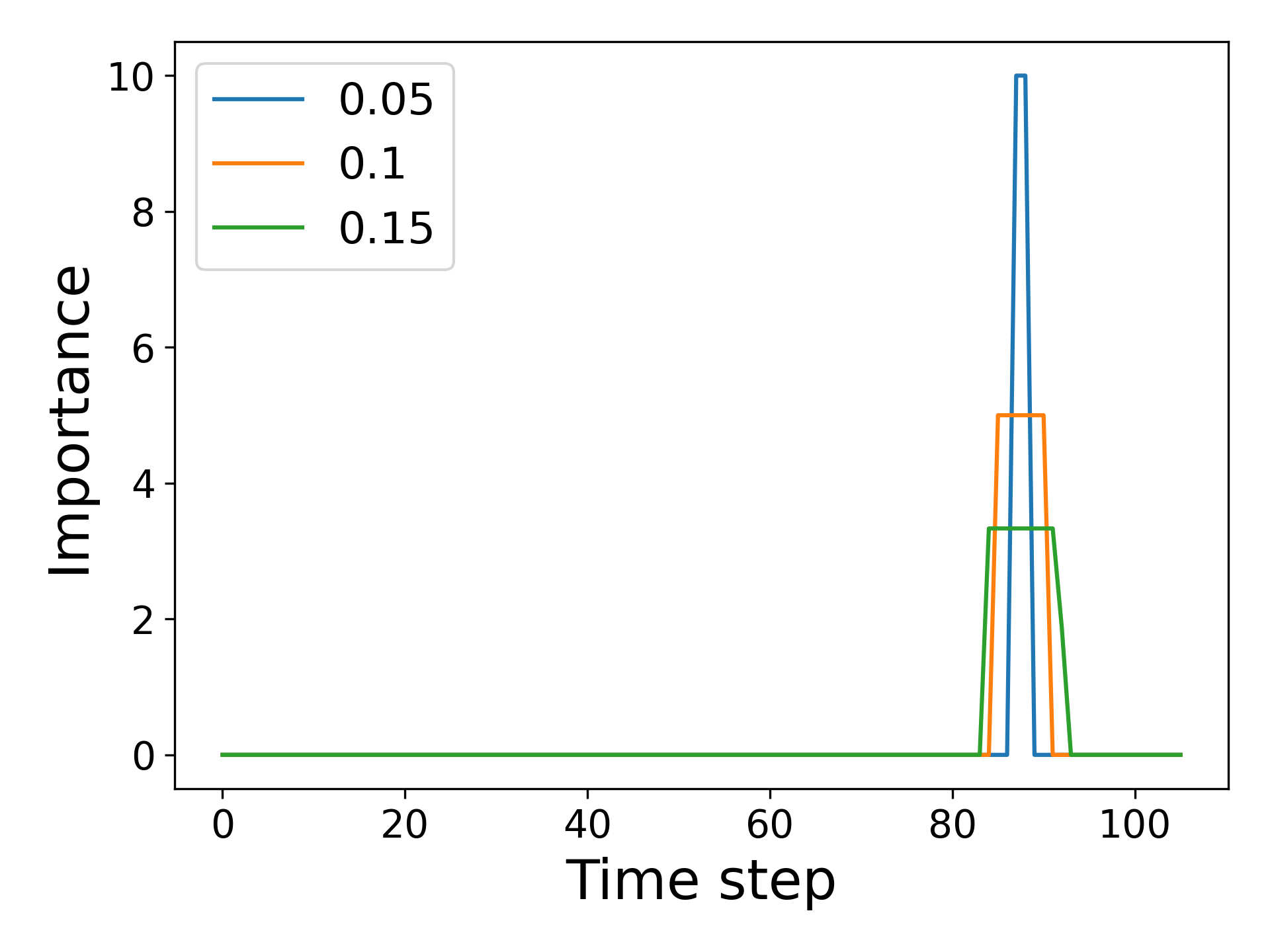}
	\caption{Importance vector of $X_t$}
	\label{gp:bang_bang_sensitivity_regular}
\end{subfigure}
\begin{subfigure}{0.45\textwidth}
\centering
	\includegraphics[width=0.75\textwidth]{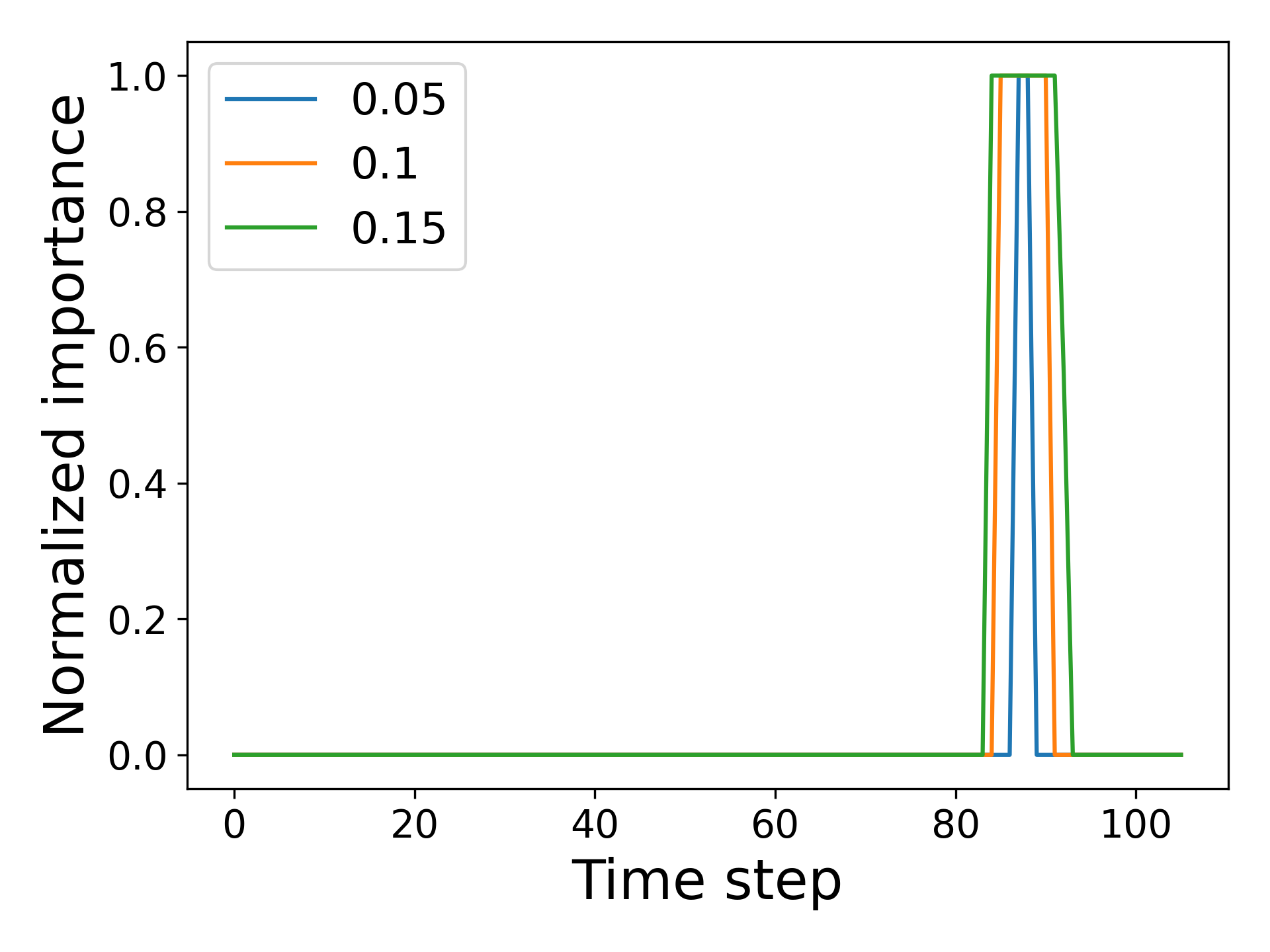}
	\caption{Normalized importance vector of $X_t$}
	\label{gp:bang_bang_sensitivity_normalized}
\end{subfigure}
\caption{Sensitivity analysis on the collision avoidance problem.}
\label{gp:bang_bang_sensitivity}

\end{figure}

Fig.~\ref{gp:bang_bang_sensitivity} shows the importance vector of $X_t$ in the collision avoidance problem and different color lines correspond to different perturbation amounts. Note that similar to the result shown in Fig.~\ref{gp:bang_importance}, the importance of $D_t$ is the same as $X_t$, and $X_{t-1}$ is the same but off by one time step. Other features have negligible importance.

There are two effects of using different perturbation amounts: 1) The number of steps with non-zero importance is increasing as $\delta$ increases since a larger $\delta$ will cause states further away from the decision boundary to cross the boundary after the perturbation; 2) The value of peak importance is lower. Since we use the action-based importance and the action is essentially binary, the difference in importance solely comes from the normalization we applied on $\delta$ (the denominator in Eq.~(\ref{eq:impor_ve}). If this is undesirable, one way to combat this is to normalize the result using the highest importance across all features and time steps. The normalized result is shown in Fig.~\ref{gp:bang_bang_sensitivity_normalized}, in which the peak value will be one regardless of $\delta$.

\subsection{Lunar Lander}

\begin{figure}[H]
\centering
\begin{subfigure}[t]{0.32\textwidth}
	\includegraphics[width=0.9\textwidth]{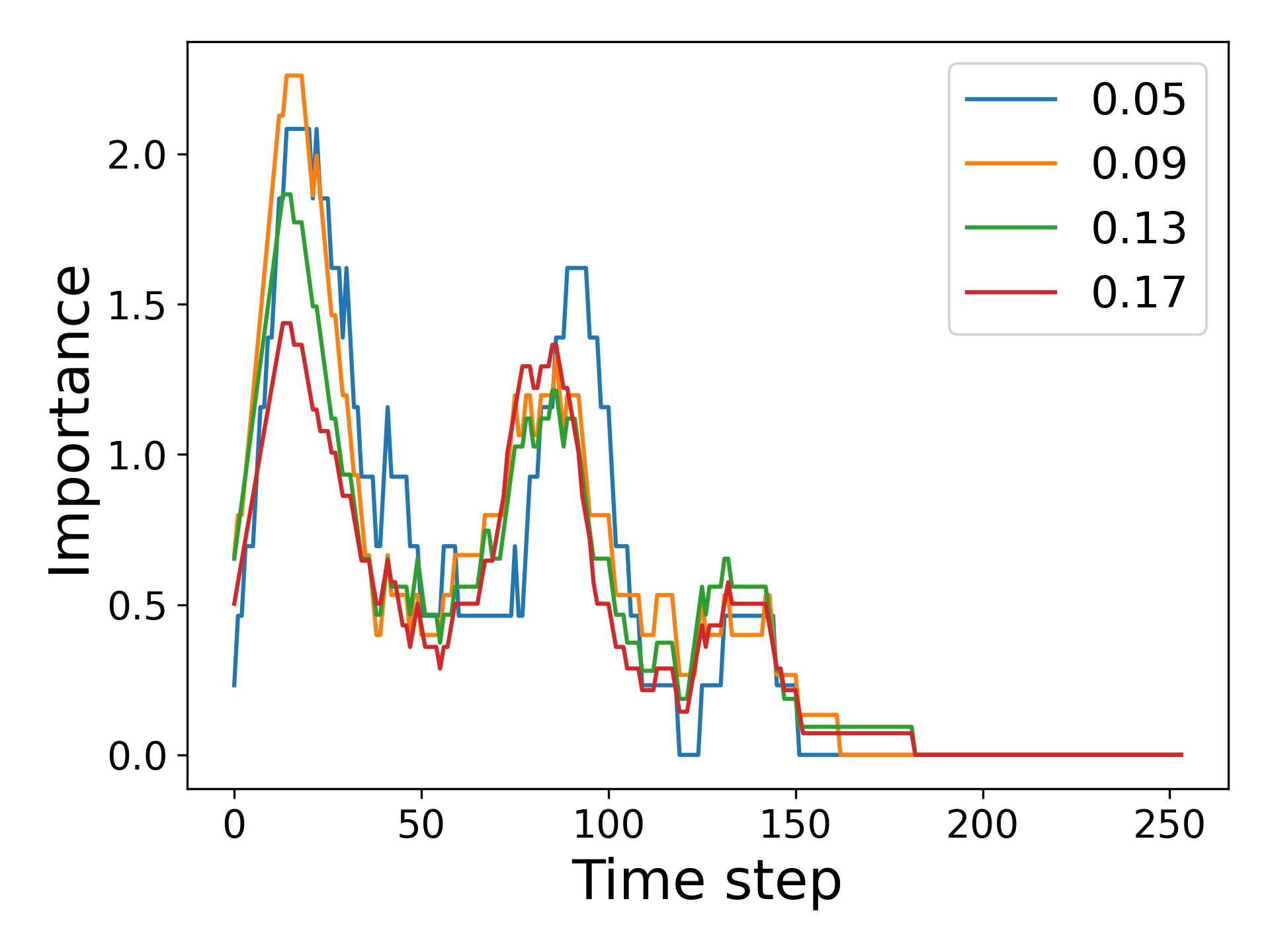}
	\caption{Importance vector of x-position}
	\label{gp:lunar_lander_sensitivity_x_pos}
\end{subfigure}
\begin{subfigure}[t]{0.32\textwidth}
	\includegraphics[width=0.9\textwidth]{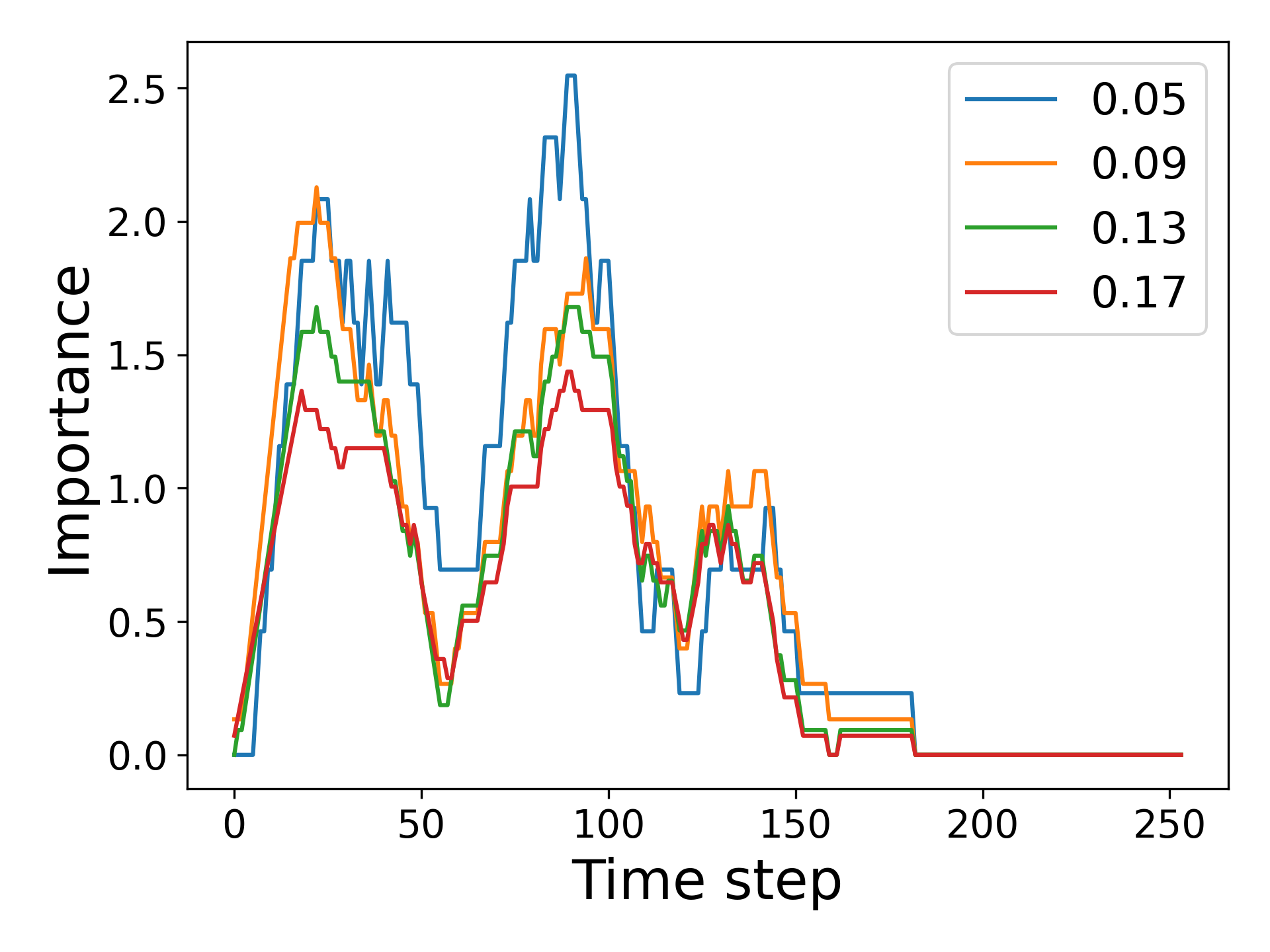}
	\caption{Importance vector of y-position}
	\label{gp:lunar_lander_sensitivity_y_pos}
\end{subfigure}
\begin{subfigure}[t]{0.32\textwidth}
	\includegraphics[width=0.9\textwidth]{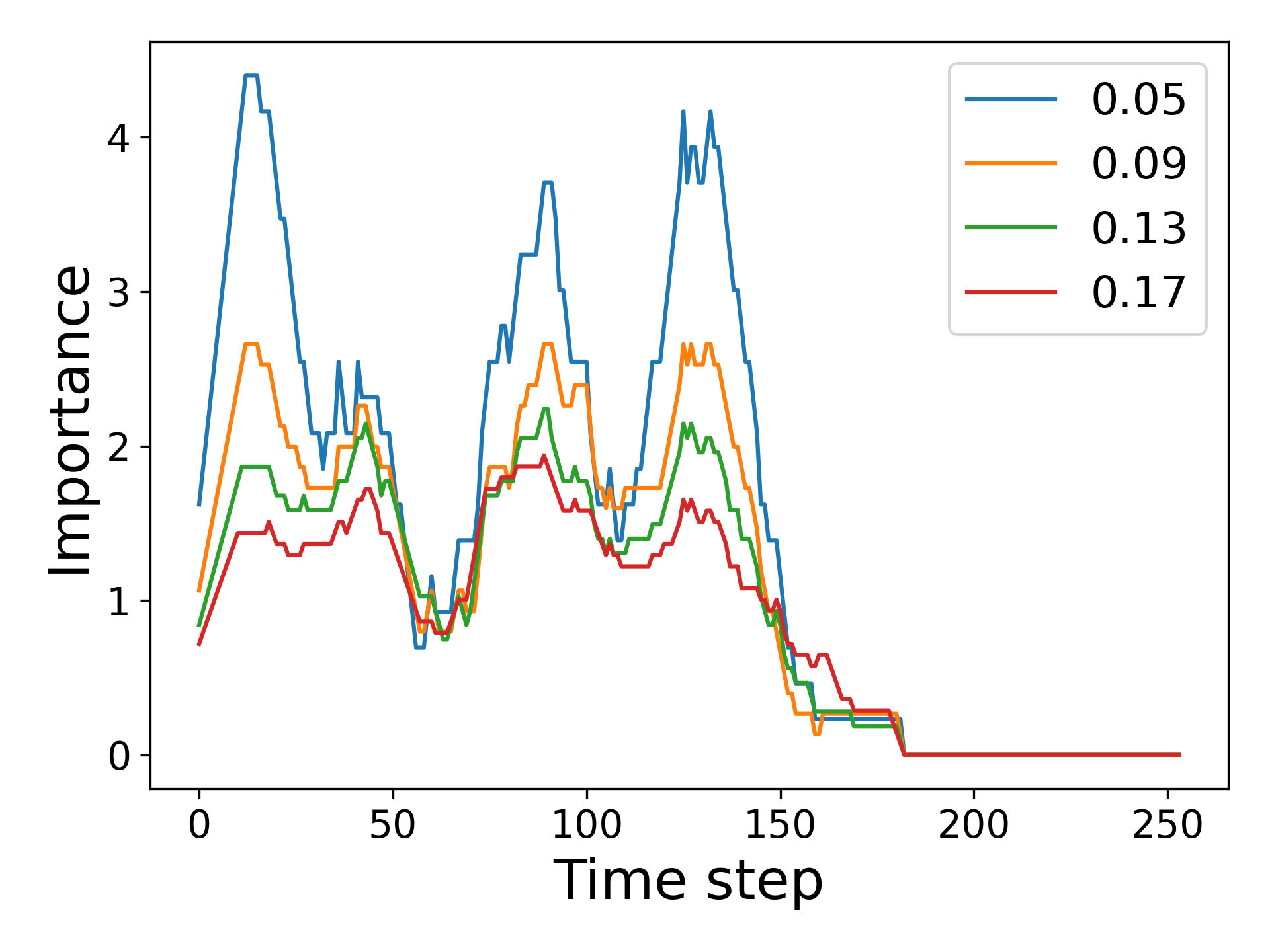}
	\caption{Importance vector of angle}
	\label{gp:lunar_lander_sensitivity_angle}
\end{subfigure}

\begin{subfigure}[t]{0.32\textwidth}
	\includegraphics[width=0.9\textwidth]{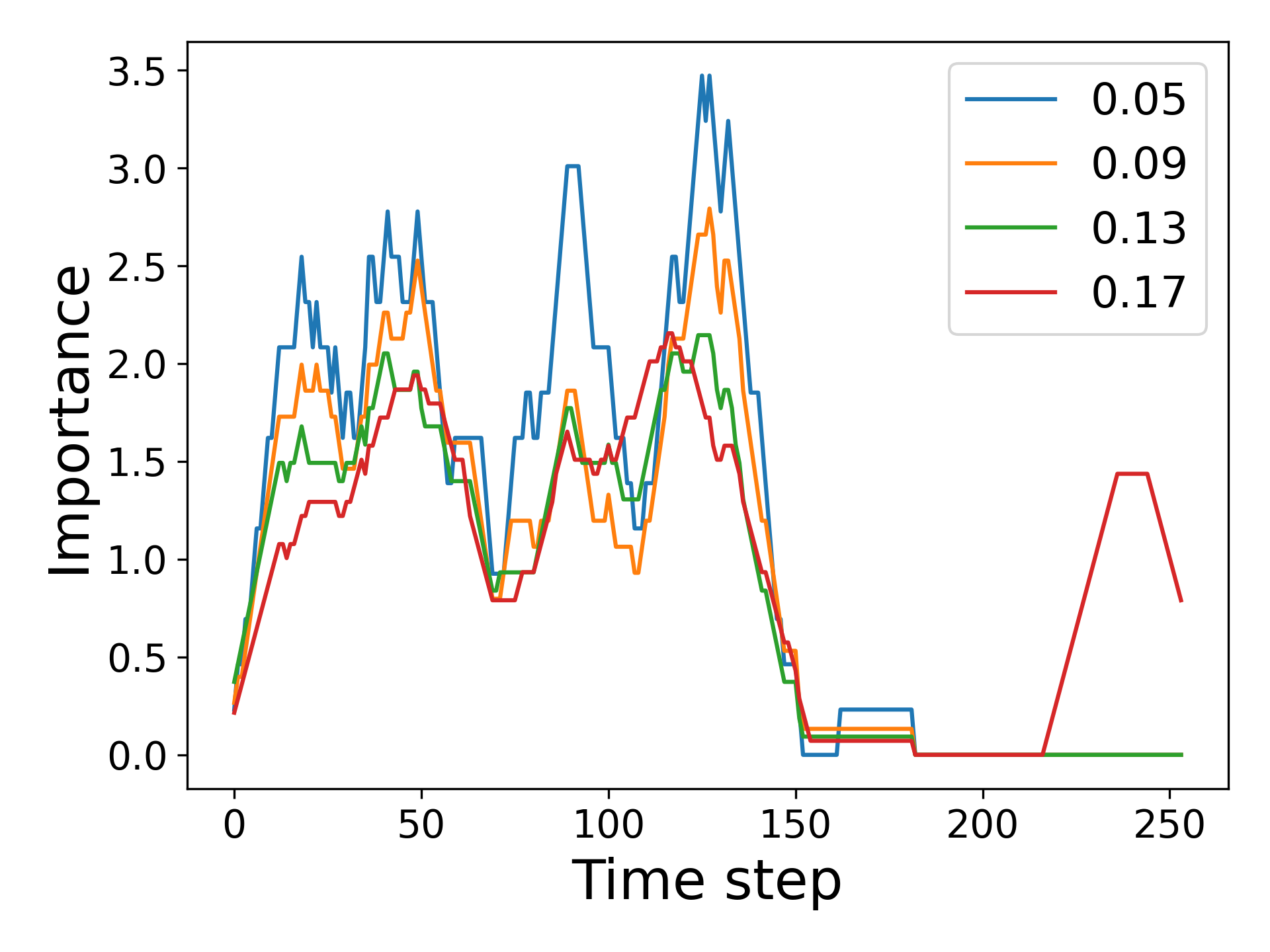}
	\caption{Importance vector of x-velocity}
	\label{gp:lunar_lander_sensitivity_x_vel}
\end{subfigure}
\begin{subfigure}[t]{0.32\textwidth}
	\includegraphics[width=0.9\textwidth]{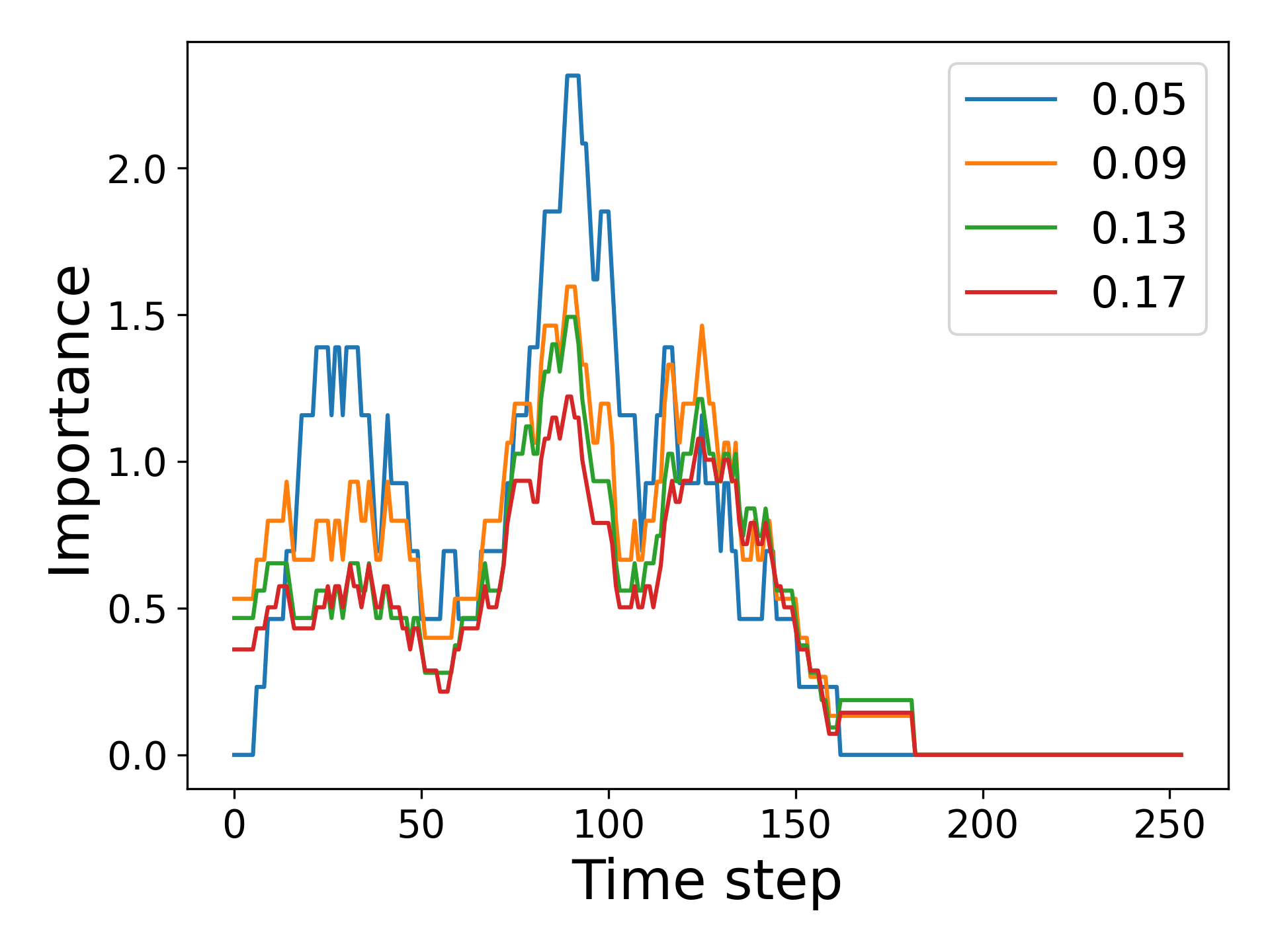}
	\caption{Importance vector of y-velocity}
	\label{gp:lunar_lander_sensitivity_x_vel}
\end{subfigure}
\begin{subfigure}[t]{0.32\textwidth}
	\includegraphics[width=0.9\textwidth]{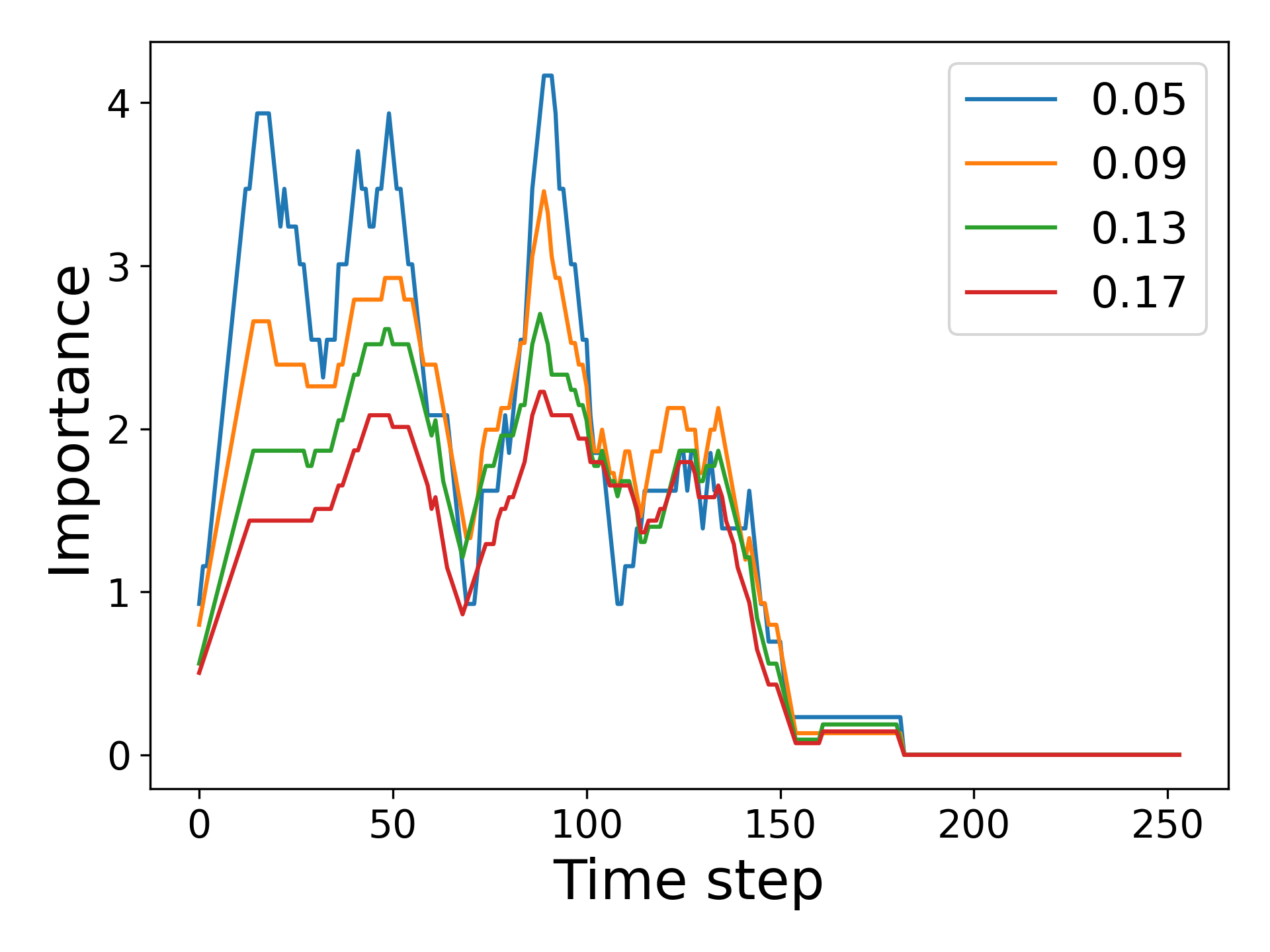}
	\caption{Importance vector of angular velocity}
	\label{gp:lunar_lander_sensitivity_angle_vel}
\end{subfigure}

\caption{Sensitivity analysis on the lunar lander environment.}
\label{gp:lunar_lander_sensitivity}

\end{figure}

Fig.~\ref{gp:lunar_lander_sensitivity} shows the sensitivity analysis on lunar lander and the different color lines correspond to different perturbation amounts. Binary features including left and right leg are not included. The general trend of the result is the same while the value and the exact shape of the curve vary slightly when different $\delta$ is used and our result is robust w.r.t. $\delta$. 

\subsection{Blackjack}
\begin{figure}[htp]
\centering
\begin{subfigure}[t]{0.45\textwidth}
	\includegraphics[width=0.9\textwidth]{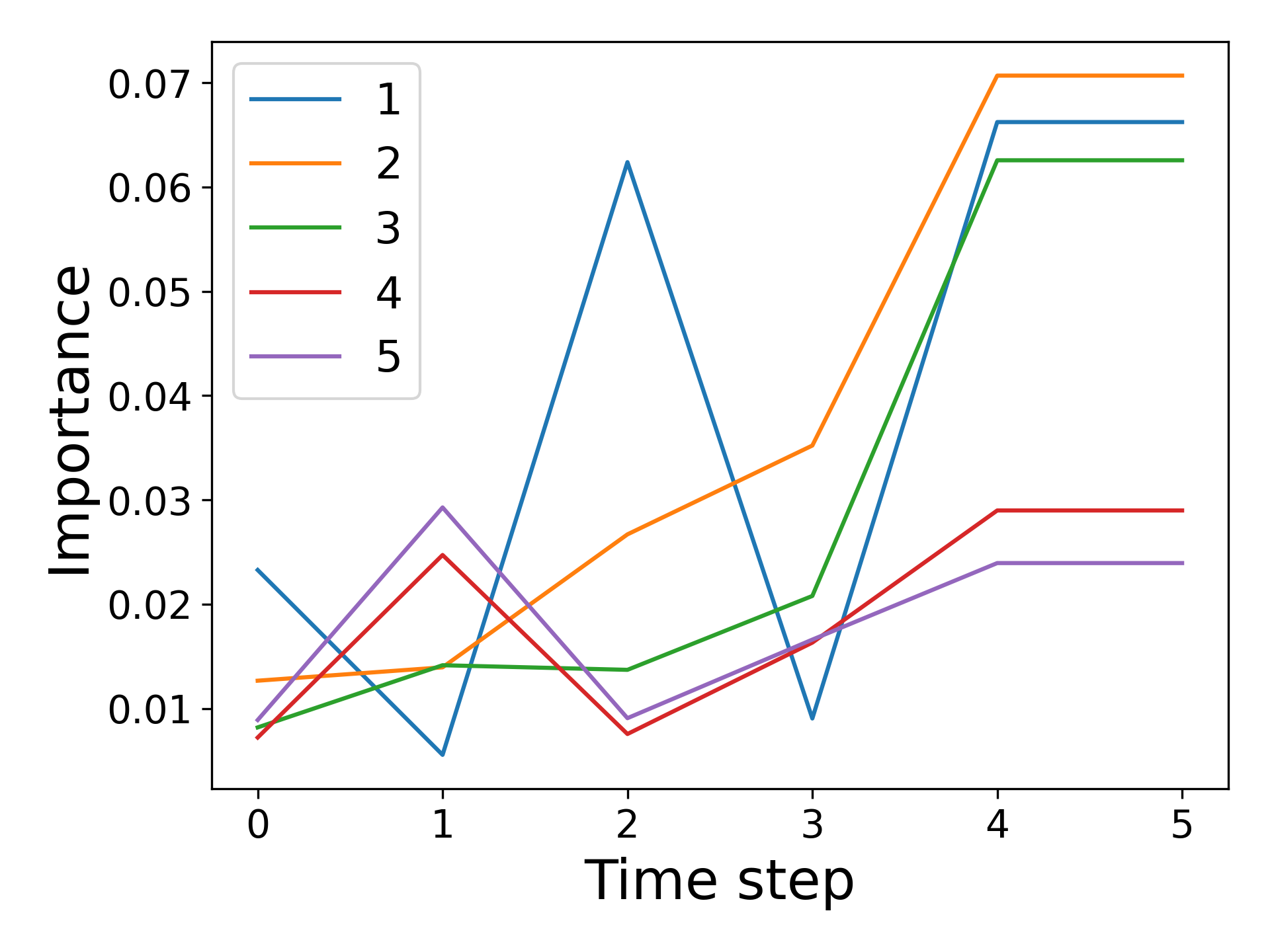}
	\caption{Importance vector of \texttt{hand}}
	\label{gp:bang_bang_sensitivity_regular}
\end{subfigure}\hfill
\begin{subfigure}[t]{0.45\textwidth}
	\includegraphics[width=0.9\textwidth]{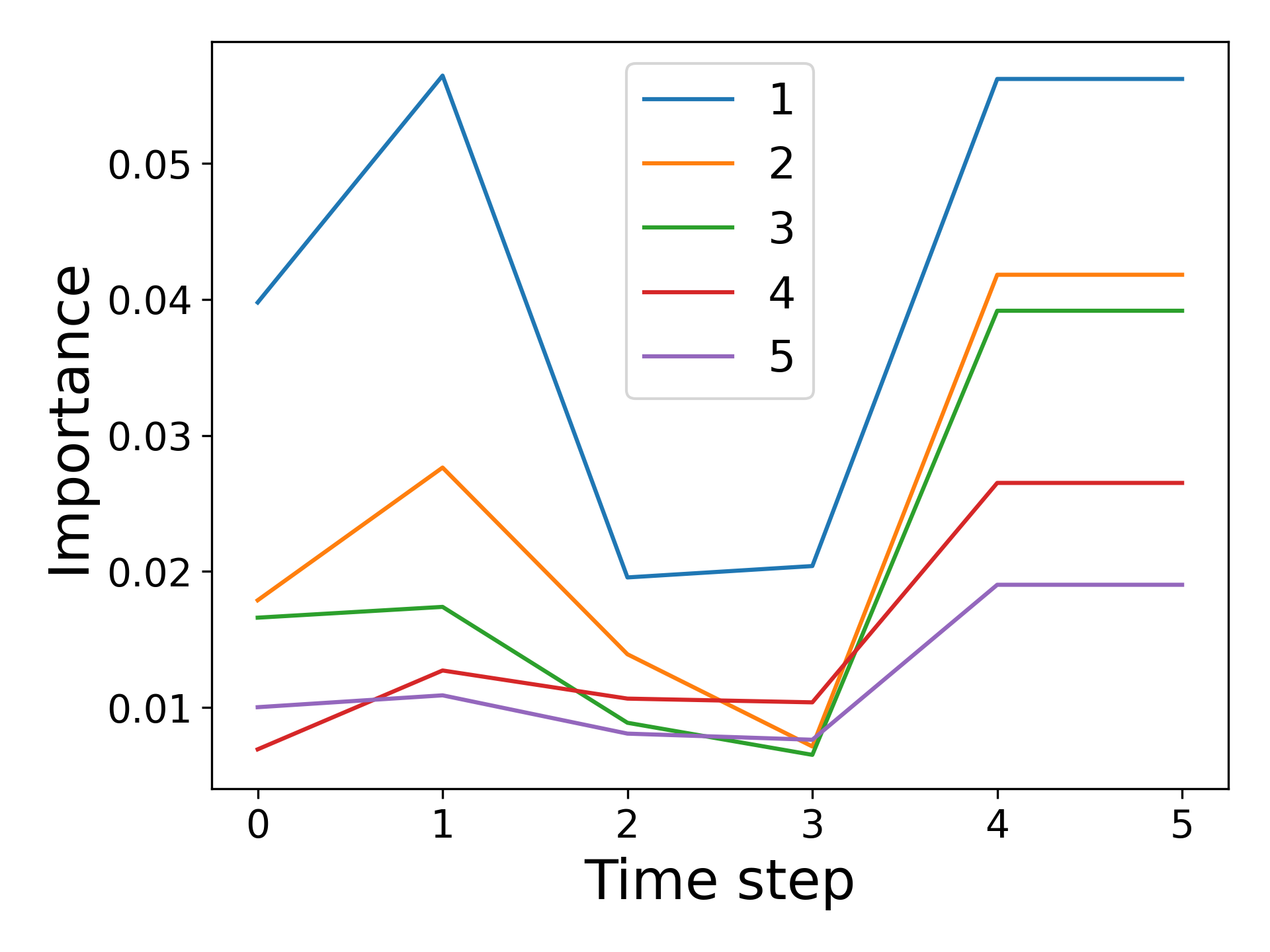}
	\caption{Importance vector of \texttt{dealer}}
	\label{gp:blackjack_sensitivity_dealer}
\end{subfigure}

\caption{Sensitivity analysis on the Blackjack environment.}
\label{gp:blackjack_sensitivity}

\end{figure}

Fig.~\ref{gp:blackjack_sensitivity} shows the sensitivity analysis for blackjack, with different color lines representing different perturbation amounts. The binary feature \texttt{ace} is not included. In blackjack, since the smallest legal perturbation amount is one and the range of the value is at most 21, increasing $\delta$ has a much larger effect on the result. However, we can observe that the general shape of the curves is similar, indicating the robustness of our method.

\section{Action-based Importance versus Q-value-based Importance}
\label{sec:compareQ}
This section discusses the comparison between the action-based importance method and the Q-value-based importance method. It demonstrates that the Q-value-based method sometimes fails to reflect the features in the state that the policy relies on.

\begin{figure}[H]
\centering
\includegraphics[width=0.4\linewidth]{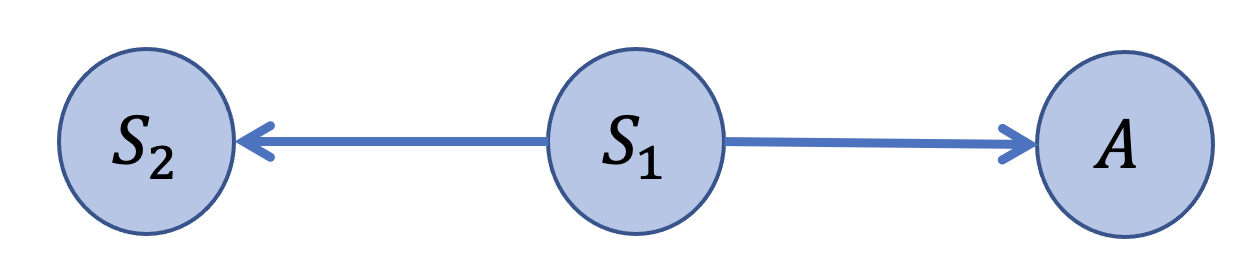}
\caption{The skeleton of SCM of the one step MDP.}
\label{gp:1step}
\end{figure}

Consider a one-step MDP with the SCM shown in Fig.~\ref{gp:1step}, where the state $\mathbf{S}=[S_1, S_2]$, $S_i \in [-1,1]$, $i=1,2$, and the action $a \in [-1,1]$. The reward is defined as $R(\mathbf{S},a)= 100\times S_2+ a \times S_1$. Under this setting, the optimal policy is:
\begin{equation*}
\label{eq:onestepp}
  A =
    \begin{cases}
      -1 & S_1 < 0\\
      1 & \text{otherwise}
    \end{cases}       
\end{equation*}
Intuitively, the policy selects the minimum value in the action space when $S_1$ is negative
, and the maximum value otherwise.

The action-based importance method correctly identifies $S_1$ as more important, as the policy only depends on $S_1$. However, the Q-value-based method produces a different result. In a one-step MDP, the Q-function is the same as the reward function. As the coefficient in the Q(reward) function is larger for $S_2$, the Q-value-based method finds $S_2$ more important, which is different from the features that the policy relies on.

\end{document}